\documentclass{article}
\usepackage{ijcai26}

\usepackage{times}
\usepackage{soul}
\usepackage{url}
\usepackage[hidelinks]{hyperref}
\usepackage[utf8]{inputenc}
\usepackage[small]{caption}
\usepackage{graphicx}
\usepackage{amsmath}
\usepackage{amsthm}
\usepackage{booktabs}
\usepackage{algorithm}
\usepackage{algorithmic}
\usepackage[switch]{lineno}

\usepackage{textcomp}
\usepackage{xcolor}
\usepackage{graphicx}
\usepackage{multirow}
\usepackage{mathtools}
\usepackage{amsmath,amssymb,amsfonts}
\usepackage{bbm}
\newtheorem{definition}{Definition}

\title{Interpretable Graph-Level Anomaly Detection via\\Contrast with Normal Prototypes}

\author{
Qiuran Zhao$^{1,2}$
\and
Kai Ming Ting$^{1,2}$\And
Xinpeng Li$^{1,2}$\\
\affiliations
$^1$State Key Laboratory for Novel Software Technology, Nanjing University, Nanjing, China\\
$^2$School of Artificial Intelligence, Nanjing University, Nanjing, China\\
\emails
zhaoqr@lamda.nju.edu.cn,
tingkm@nju.edu.cn,
lixp@lamda.nju.edu.cn
}

\begin{document}

\maketitle

\begin{abstract}
The task of graph-level anomaly detection (GLAD) is to identify anomalous graphs that deviate significantly from the majority of graphs in a dataset.
While deep GLAD methods have shown promising performance, their black-box nature limits their reliability and deployment in real-world applications. Although some recent methods have made attempts to provide explanations for anomaly detection results, they either provide explanations without referencing normal graphs, or rely on abstract latent vectors as prototypes rather than concrete graphs from the dataset.
To address these limitations, we propose \textbf{Prototype-based Graph-Level Anomaly Detection (ProtoGLAD)}, an interpretable unsupervised framework that provides explanation for each detected anomaly by explicitly contrasting with its nearest normal prototype graph.
It employs a point-set kernel to iteratively discover multiple normal prototype graphs and their associated clusters from the dataset, then identifying graphs distant from all discovered normal clusters as anomalies.
Extensive experiments on multiple real-world datasets demonstrate that ProtoGLAD achieves competitive anomaly detection performance compared to state-of-the-art GLAD methods while providing better human-interpretable prototype-based explanations.
\end{abstract}

\section{Introduction}
Graphs are data structures widely used to represent complex relationships in various domains such as chemistry, traffic systems, and social networks. Recently, graph-level anomaly detection (GLAD) has emerged as a critical but challenging task in the graph data mining community \cite{luo2022deep,qiu2022raising,zhao2023using,ma2022deep,liu2023towards,kim2024rethinking}.
GLAD aims to identify graphs that deviate significantly from the majority of graphs in a dataset of graphs.
Reliable GLAD methods enable a variety of applications, including the detection of toxic molecular compounds, finding malicious communication patterns in networks, and identifying abnormal brain connectivity profiles in neuroscience.

However, despite their promising detection performance, most advanced GLAD methods operate as black-box models.
They detect anomalies based on anomaly scores derived from, for example, the reconstruction error of a graph autoencoder \cite{luo2022deep}, the learning discrepancy from a knowledge distillation framework \cite{ma2022deep}, or the consistency score from cross-view mutual information maximization \cite{liu2023towards}.
The anomaly scores offered by these methods are self-referential, which means they measure how well a graph conforms to a compressed version of itself or to an augmented view of itself, rather than comparing it to normality.
Consequently, it is unclear how the detected anomalies can be understood when the detectors cannot provide human-interpretable explanations or verify whether their detected anomalies are indeed rare and different from the majority of the dataset.

Recently, some works \cite{liu2023towards,yang2025gladpro,zhang2025x} have attempted to provide explanations for their detected graph anomalies. However, existing approaches are limited as they either fail to provide a comparative normal reference to verify why a detected anomaly is deviant, or use learnable latent vectors as prototypes, which may deviate from real normal graphs and lead to unreliable normal references for detection and explanation.

To address these issues, we propose Prototype-based Graph-Level Anomaly Detection (ProtoGLAD).
We argue that this paradigm is particularly well suited for GLAD, as it discovers multiple prototype graphs and their associated normal clusters, which together serve as normal references for anomaly detection and explanation.
The anomaly score of a graph is then determined by its similarity to the nearest normal cluster.
This design naturally aligns with the definition of graph anomalies as rare deviations from normal distributions and provides an inherent basis for explainability, as each detected anomaly can be directly contrasted with its nearest normal prototype. 

The main contributions of this work are:
\begin{itemize}
    \item Proposing an effective, interpretable GLAD framework capable of detecting anomalous graphs accurately based on their similarity with respect to a cluster of normal graphs discovered from the dataset.
    \item Developing a node-level explanation mechanism that uses a similarity measure between a node and a graph, enabling the identification of nodes and substructures that are responsible for the anomalousness of a detected graph.
    \item Conducting extensive experiments on multiple real-world datasets, demonstrating that ProtoGLAD achieves competitive anomaly detection performance compared to state-of-the-art GLAD methods while providing clear and interpretable prototype-based explanations.
\end{itemize}

\section{Related Work}
\label{sec:RW}
In this section, we briefly review related work.

\subsection{Graph-level anomaly detection}

Graph-level anomaly detection (GLAD) aims to identify anomalous graphs that exhibit significant deviations from the majority of graphs in a dataset. 
Existing GLAD methods can be broadly categorized into two types:
\textbf{two-step} methods and \textbf{end-to-end} methods.
\textbf{Two-step methods} first generate graph-level embeddings using graph kernels \cite{vishwanathan2010graph,neumann2016propagation} or self-supervised representation learning (\textit{e.g.}, Graph2Vec \cite{narayanan2017graph2vec}, InfoGraph \cite{sun2019infograph}), followed by applying an anomaly detector such as OCSVM  \cite{amer2013enhancing}, LOF \cite{breunig2000lof}, or Isolation Forest \cite{liu2008isolation}.
\textbf{End-to-end methods}, on the other hand, leverage GNNs \cite{kipf2016semi,xu2018powerful} to jointly learn graph representations and detect anomalies in a unified framework.
Among them, some approaches presume that the anomaly labels of the graphs are accessible during model training, and thus frame GLAD as a supervised classification problem \cite{zhang2022dual,ma2023towards,dong2023rayleigh}.
However, their strong dependence on labeled anomalies renders their application difficult in a real-world scenarios where such annotations are few or, more often than not, nonexistent.
Therefore, most approaches focus on unsupervised learning, where only normal samples are used during the training phase.
For example, some of them adopt GNN-based graph reconstruction \cite{luo2022deep,kim2024rethinking}, where anomalies are identified as graphs that are poorly reconstructed by the model, while some others like GLocalKD \cite{ma2022deep} uses a knowledge distillation framework where anomalies are detected as those having large knowledge discrepancies.

\subsection{Explainable GLAD}
Explainable graph-level anomaly detection has attracted increasing attention, and several recent methods have attempted to provide explanations for their detected anomalies.
SIGNET \cite{liu2023towards} identifies the specific substructures within a single graph that contribute to its anomaly score. It is done by extracting an important ``bottleneck'' subgraph using the Information Bottleneck principle.
While it highlights the suspicious subgraph of the detected anomaly, it fails to provide a contrastive reference to a normal counterpart, making it difficult to verify why the highlighted subgraph is considered deviant.

GLADPro \cite{yang2025gladpro}, aims to provide a general understanding of the model's behavior by learning global prototypes. It extends the IB principle to learn a set of prototypes that capture important subgraph patterns across the entire dataset.
However, a key limitation is that these prototypes are modeled as abstract, learnable vectors in a latent space, rather than actual graphs directly discovered from the dataset, and interpreting a high-dimensional latent vector is a challenging task.

More recently, X-GAD \cite{zhang2025x} aggregates common structural and attribute characteristics across anomalous graphs to construct \textbf{anomaly prototypes}.
By contrasting graphs with these learned anomaly prototypes, X-GAD provides explanations that highlight recurring anomalous patterns shared among abnormal graphs.
However, X-GAD relies on the availability of anomaly labels during training, which are often expensive or inaccessible.
Furthermore, the presupposition of anomaly prototypes is a restrictive assumption, as anomalies do not necessarily exhibit consistent patterns.
These factors constrain the practical utility of X-GAD in real-world applications.

\section{Preliminaries}
\subsection{Problem definition}

Let $M=\{\mathcal{G}_i|i=1,\dots,n\}$ be a set of  $n$ attributed graphs, where each graph $\mathcal{G}_i=(\mathcal{V}_i, \mathcal{E}_i)$ has a set of nodes $\mathcal{V}_i$ and a set of edges $\mathcal{E}_i$.
Each node $v\in \mathcal{V}$ is associated with an attribute vector $\mathbf{x}_v\in \mathbb{R}^m$ with $m$ attributes. We assume the majority of graphs in $M$ are normal and can be grouped into $k$ distinct normal clusters $C = \{C_1, ..., C_k\}$, where each cluster has a prototype $P_i$.
The anomalous graphs are few and dissimilar from the majority of the graphs in $M$.
This process requires a similarity measure $sim(\cdot, \cdot)$ that can compute the similarity between a single graph and a \textit{set} (or distribution) of graphs.

\begin{definition}[\textbf{Graph-Level Anomaly Detection}]
Graph-level anomaly detection aims to identify anomalous graphs which have characteristics different from those of the \textbf{majority} of the graphs in the given set $M$ of graphs, and to rank all the graphs on the basis of their anomaly scores such that the anomalous graphs are ranked higher than the normal ones.
\end{definition}

\begin{definition}[\textbf{Normal and Anomalous Graphs}]
Given $M=\{\mathcal{G}_i\}$, a set of $k$ normal clusters $C = \{C_j\}_{j=1}^k$ discovered from $M$, and a point-set similarity function $sim(\mathcal{G}, C_j)$ that measures the similarity between a graph $\mathcal{G}$ and the distribution of cluster $C_j$:
\begin{itemize}
    \item A graph $\mathcal{G}^N$ is \textbf{normal} if it belongs to a cluster $C_j$ and is close to that cluster's overall distribution.
    \item A graph $\mathcal{G}^A$ is \textbf{anomalous} if it does not belong to any cluster and is dissimilar to all normal cluster distributions.
\end{itemize}
and they have the following relationship:
\[ \max_{j} \left( sim(\mathcal{G}^N, C_j) \right) \gg \max_{j} \left( sim(\mathcal{G}^A, C_j) \right) \]
\label{def-anomaly}
\end{definition}

\subsection{Weisfeiler-Lehman graph kernels}

Graph kernels are a class of methods for measuring the similarity between graphs.
Most existing graph kernels are based on the R-Convolution kernel \cite{haussler1999convolution}, which operates by extracting substructures from graphs and representing these substructures in a vector form.
The kernel between two graphs is then computed based on the similarity between these substructures.
This process of converting graphs into vector representations is often referred to as embedding.

One of the most influential embedding techniques is the Weisfeiler-Lehman (WL) scheme  \cite{leman1968reduction,shervashidze2011weisfeiler}, which captures the dependencies in a graph in the form of subtrees.
The WL scheme applies specifically to labeled graphs.
Over time, there have been several extensions and variations of the WL scheme  \cite{shervashidze2011weisfeiler,morris2017glocalized}.
For instance, a different version of the WL scheme has been proposed to handle attributed graphs, \textit{i.e.}, graphs with continuous node features and possibly weighted edges  \cite{togninalli2019wasserstein}.
In this approach, each graph is encoded as a histogram that reflects the distribution of node embeddings produced by the WL process. 

Let $\mathcal{G} = (\mathcal{V}, \mathcal{E})$ be an attributed graph and $\mathcal{N}(v)$ be the set of points in the 1-hop neighborhood of $v$, excluding $v$.

\medskip

\begin{definition} A Weisfeiler-Lehman (WL) scheme \cite{togninalli2019wasserstein} that embeds node $v$ associated with $\mathcal{G}$, denoted as $\mathcal{G}(v)$, is defined as:
\[
\mathbf{x}_v^{(h)} = [\mathbf{x}_v^0, \mathbf{x}_v^1, \dots, \mathbf{x}_v^h]
\]
where $\mathbf{x}_v^h$ represents the embedding of $\mathcal{G}(v)$ at iteration $h$:
\[
\mathbf{x}_v^h = \frac{1}{2} \left( \mathbf{x}_v^{(h-1)} + \frac{1}{|\mathcal{N}(v)|} \sum_{u \in \mathcal{N}(v)} \mathbf{x}_u^{(h-1)} \right)
\]
\end{definition}
\medskip


\subsection{Isolation Kernel}

Isolation Kernel (IK) is a data-dependent kernel that derives directly from data without explicit learning, and it has no closed-form expression \cite{ting2018isolation}.
It has been shown to improve the task-specific performance of SVM \cite{ting2018isolation} and density-based clustering \cite{qin2019nearest} simply by replacing the data-independent kernel or distance in the algorithms. 

Let $D = \bigcup_{v \in \mathcal{V}_i, i \in [1,n]} \mathbf{x}_v$ be the set of all node vectors in the given dataset $M$ of graphs.
Here we use IK to map each node vector $\mathbf{x}_v \in \mathbb{R}^m$ into a vector $\varphi(\mathbf{\mathbf{x}_v} \mid D) \in \{0, 1\}^{t \times \psi}$ as is done in \cite{xu2021isolation}, where the use of IK has been shown to be the key to improving the discriminative power of the WL scheme.

Let $\mathbb{H}_\psi(D)$ denote the set of all partitionings $H$ admissible from $\mathcal{D} \subset D$, where each point $\hat{\mathbf{x}}_v \in \mathcal{D}$ has equal probability of being selected from $D$ and $|\mathcal{D}| = \psi$. Each $\theta[\hat{\mathbf{x}}_v] \in H$ isolates a point $\hat{\mathbf{x}}_v \in \mathcal{D}$. 

\begin{definition}[ \cite{ting2018isolation,qin2019nearest}]
Isolation Kernel of $\mathbf{x}_1$ and $\mathbf{x}_2$ is defined to be the expectation taken over the distribution of partitionings $H \in \mathbb{H}_\psi(D)$ that both $\mathbf{x}_1$ and $\mathbf{x}_2$ fall into the same isolating partition $\theta[\hat{\mathbf{x}}_v] \in H$, where $\hat{\mathbf{x}}_v \in \mathcal{D} \subset D$, $\psi = |\mathcal{D}|$:
\begin{equation*}
\begin{aligned}
\kappa_\psi(\mathbf{x}_1, \mathbf{x}_2 \mid D) 
&= \mathbb{E}_{\mathbb{H}_\psi(D)} \left[ \mathbbm{1}(\mathbf{x}_1, \mathbf{x}_2 \in \theta \mid \theta \in H) \right] \\
& \approx \frac{1}{t} \sum_{i=1}^t \mathbbm{1}(\mathbf{x}_1, \mathbf{x}_2 \in \theta \mid \theta \in H_i) \\
&= \frac{1}{t} \sum_{i=1}^t \sum_{\theta \in H_i} \mathbbm{1}(\mathbf{x}_1 \in \theta)\mathbbm{1}(\mathbf{x}_2 \in \theta)
\end{aligned}
\label{eq:IK}
\end{equation*}

\end{definition}

where $\mathbbm{1}(\cdot)$ is an indicator function;
$\kappa_\psi$ is constructed using a finite number of partitionings $H_i$, $i = 1, \dots, t$, where each $H_i$ is created using randomly subsampled $\mathcal{D}_i \subset D$;
and $\theta$ is a shorthand for $\theta[\hat{\mathbf{x}}_v]$.

\begin{definition}[\textbf{Feature map of Isolation Kernel.} \cite{ting2020isolation}]
For node vector $\mathbf{x}_v \in \mathbb{R}^m$, the feature mapping $\varphi: \mathbb{R}^m \rightarrow \{0, 1\}^{t \times \psi}$ of $\kappa_\psi$ maps $\mathbf{x}_v$ into a vector that represents the partitions in all the partitionings $H_i \in \mathbb{H}_\psi(D)$, $i = 1, \dots, t$, where $\mathbf{x}_v$ falls into either one of the $\psi$ hyperspheres or none in each partitioning $H_i$.
\label{def-feature-map}
\end{definition}

Rewriting Equation~\ref{eq:IK} using $\varphi$ gives:

\[
\kappa_\psi(\mathbf{x}_1, \mathbf{x}_2 \mid D) = \frac{1}{t} \langle \varphi(\mathbf{x}_1 \mid D), \varphi(\mathbf{x}_2 \mid D) \rangle
\]

\section{Proposed framework: ProtoGLAD}


\subsection{Graph embedding}

The Weisfeiler-Lehman (WL) embedding used here is the same as that introduced in section~\ref{sec:RW}.
The only exception is that we use the IK mapped vector $\varphi(\mathbf{v})$ rather than node vector $\mathbf{x}_v$ as the input representation of the WL scheme, which is the same as in \cite{xu2021isolation}.

Let $\varphi^0(\mathbf{x}_v) = \varphi(\mathbf{x}_v)$ be the IK mapped vector of node $\mathbf{x}_v$ for a node $v$ in a graph $\mathcal{G} = (\mathcal{V}, \mathcal{E})$. The WL embedding at iteration $h > 0$ is recursively defined as:
\begin{equation*}
    \varphi^{h}(\mathbf{x}_v) = \frac{1}{2} \left( \varphi^{h-1}(\mathbf{x}_v) + \frac{1}{|\mathcal{N}(v)|} \sum_{u \in \mathcal{N}(v)} \varphi^{h-1}(\mathbf{x}_u) \right)
\label{eq-WL}
\end{equation*}

We view each graph $\mathcal{G} = (\mathcal{V}, \mathcal{E})$ as having a representative sample of WL-embedded node vectors $\varphi^{h}(\mathbf{x}_v)$ ($\forall v\in\mathcal{V}$) of an unknown distribution, where each node vector is represented using the IK-WL embedding.
Then, a mean embedding can be used to represent each graph.

For a graph $\mathcal{G} = (\mathcal{V}, \mathcal{E})$, its  embedding is given as the mean embeddings of all nodes in $\mathcal{G}$:
\[
\Phi(\mathcal{G}) = \frac{1}{|\mathcal{V}|} \sum_{v \in \mathcal{V}} \varphi^h(\mathbf{x}_v).
\]

The similarity between any pair of two graphs is calculated by a dot product between the two mean embeddings or feature mean maps, \textit{i.e.},
\[
sim(\mathcal{G}, \mathcal{G}') = \langle {\Phi}(\mathcal{G}),\ {\Phi}(\mathcal{G}') \rangle
\]

\subsection{Anomalous graph detection}

Once graph-level embeddings are obtained for each graph in the dataset, the GLAD problem can be naturally reformulated as a point anomaly detection task in the embedding space.
Each graph $\mathcal{G}_i$ is represented by a fixed-dimensional vector $\Phi(\mathcal{G}_i)$ that captures its structural and semantic properties.
Then we utilize a point-set kernel $\hat{K}$ to measure the similarity between a single graph and a set of graphs. 

This point-set kernel is defined as:

$$ \hat{K}(x, C) = \langle\Phi(x), \hat{\Phi}(C)\rangle $$
and

$$ \hat{\Phi}(C) = \frac{1}{|C|}\sum_{y \in C}\Phi(y) $$
where $\hat{\Phi}$ is the kernel mean map of $\hat{K}$; $\Phi$ is the feature map of a point-to-point kernel $\kappa$; and $\langle a, b \rangle$ denotes a dot product between two vectors $a$ and $b$. 
The point-to-point kernel, expressed in terms of a dot product, is given as follows:
$$ \kappa(x, y) = \langle\Phi(x), \Phi(y)\rangle$$
In this paper, follow a prior work  \cite{ting2022point}, we propose to use Isolation Kernel introduced above as $\kappa$ in $\hat{K}$ because it has a finite-dimensional feature map, and its similarity adapts to local density of the data distribution of a given dataset.

ProtoGLAD employs the point-set kernel $\hat{K}$ to discover clusters by first locating a high-density data point which is treated as the prototype for that cluster.
A cluster is then ``grown'' from this prototype by absorbing all points that are sufficiently similar to the cluster, based on a growth rate $\rho$.
The process repeats for subsequent clusters using the unassigned points in $\Pi$.
It continues until $\Pi$ is empty or no remaining point has a similarity greater than a threshold $\tau$.
The final anomaly score for any graph $G_i$ is its point-set similarity to its nearest cluster.
Graphs dissimilar to all clusters are thus identified as anomalies.
The detailed algorithm is given in Algorithm~\ref{alg:ProtoGLAD}.

\begin{algorithm}[t]
\caption{\textbf{ProtoGLAD}: Prototype-based Graph-Level Anomaly Detection}
\label{alg:ProtoGLAD}
\begin{algorithmic}[1]
\REQUIRE Dataset of graphs $M = \{\mathcal{G}_i=(\mathcal{V}_i, \mathcal{E}_i)| i = 1, \ldots, n\}$, $\boldsymbol{\tau}$: similarity threshold, $\varrho$: growth rate
\ENSURE Anomaly scores $S=\{s_i\}_{i=1}^n$ for the graphs in $M$, a set of prototype graphs $P = \{\mathcal{G}_p^j\}_{j=1}^k$. 
\STATE Obtain graph embedding $g_i$ for each graph $\mathcal{G}_i$ in $M$, $G_{all} = \{g_i, i = 1, \ldots, n\}$ 
\STATE $\Pi = G_{all}$
\STATE $k=0$
\WHILE {$|\Pi| > 1$}
    \STATE $g_p = \arg \max_{g \in \Pi} \hat{K}(g, \Pi)$
    \STATE $g_q = \arg \max_{g \in \Pi \setminus \{g_p\}} \hat{K}(g, \{g_p\})$
    \STATE $\gamma = (1 - \varrho) \times \hat{K}(g_q, \{g_p\})$
    \IF {$\gamma \le \boldsymbol{\tau}$}
        \STATE \textbf{break}
    \ENDIF
    \STATE $k=k+1$
    \STATE $\mathcal{G}_p^k = g_p$
    \STATE $C^k = \{g_p, g_q\}$
    \WHILE {$\gamma > \boldsymbol{\tau}$}
        \STATE $C_{\text{new}}^k = \{g \in \Pi \mid \hat{K}(g, C^k) > \gamma\}$
        \STATE $C^k = C_{\text{new}}^k$
        \STATE $\gamma = (1 - \varrho)\gamma$
    \ENDWHILE
    
    \STATE $\Pi = \Pi \setminus C^k$
\ENDWHILE
\FOR {$i = 1$ to $n$}
    \STATE $s_i = \max_{j=1, \dots, k} \hat{K}(g_i, C^j)$
\ENDFOR
\STATE \textbf{return} $S = \{s_i\}_{i=1}^n$, $P = \{\mathcal{G}_p^j\}_{j=1}^k$
\end{algorithmic}
\end{algorithm}

\subsection{Explain why a detected graph is anomalous}
Once the anomaly score is obtained, for a detected anomalous graph $\mathcal{G}_A$, we identify its most similar prototype $\mathcal{G}_p^*$.
Then, we explain the anomalousness of $\mathcal{G}_A$ by assigning each node $u\in\mathcal{V}_A$ a score measuring how well it aligns with the nearest prototype.
We define the node normality score as the inner-product similarity between the node embedding and the nearest prototype embedding:
\begin{equation*}
c(u)
= \langle \varphi^h(u), \Phi(\mathcal{G}_p^*)\rangle .
\end{equation*}
Intuitively, a lower $c(u)$ indicates that node $u$ contributes little to the similarity between $\mathcal{G}_A$ and its nearest normal prototype, signifying the contribution of this node to the anomalousness of $\mathcal{G}_A$.

In the same way, scores for nodes in $\mathcal{G}_p^*$ can be computed with respect to $\mathcal{G}_A$.
This identifies the lowest scored nodes in $\mathcal{G}_p^*$ that are least similar to the nodes in $\mathcal{G}_A$. 
Thus, the pair-sets of lowest scored nodes in $\mathcal{G}_A$ and $\mathcal{G}_p^*$ explain how the detected anomalous graph differs from its most similar prototype. 



\section{Experiments}

\subsection{Experimental Setup}




\textbf{Datasets.}
We employ eight publicly available real-world datasets from the TUDataset benchmark \cite{Morris+2020}, which are collected from various critical domains including bioinformatics and social networks. 

\textbf{Baselines.}
Six unsupervised GLAD methods are selected as baseline methods:
\begin{itemize}
    \item \textbf{Two-step Method}: We choose \textbf{WL-iForest}, a classic approach that combines the Weisfeiler-Lehman graph kernel \cite{shervashidze2011weisfeiler} for graph embedding with the Isolation Forest \cite{liu2008isolation} for anomaly detection.
    \item \textbf{Deep Learning Methods}: We include five state-of-the-art deep GLAD methods: \textbf{GLADC} \cite{luo2022deep}, \textbf{GLocalKD} \cite{ma2022deep}, \textbf{OCGTL} \cite{qiu2022raising}, \textbf{SIGNET} \cite{liu2023towards}, and \textbf{GLADPro} \cite{yang2025gladpro}.
\end{itemize}
It is important to note that among these baselines, only \textbf{SIGNET} and \textbf{GLADPro} are capable of providing explanations for their detection results.


\subsection{Anomaly Detection Performance}
\begin{table*}[!t]
\centering
\begin{tabular}{l|c|c|c|c|c|c|c}
\hline
\textbf{Dataset} & \textbf{ProtoGLAD} & \textbf{GLADC} & \textbf{GLocalKD} & \textbf{SIGNET} & \textbf{WL-iForest} & \textbf{OCGTL} & \textbf{GLADPro} \\ \hline
AIDS & \textbf{0.980$\pm$0.005} & 0.967$\pm$0.005 & 0.969$\pm$0.012 & 0.976$\pm$0.009 & 0.614$\pm$0.031 & 0.975$\pm$0.021 & 0.526$\pm$0.148 \\
BZR & \textbf{0.856$\pm$0.005} & 0.678$\pm$0.032 & 0.681$\pm$0.056 & 0.779$\pm$0.069 & 0.528$\pm$0.023 & 0.519$\pm$0.031 & 0.606$\pm$0.095 \\
COX2 & \textbf{0.709$\pm$0.015} & 0.615$\pm$0.044 & 0.602$\pm$0.059 & 0.645$\pm$0.078 & 0.512$\pm$0.033 & 0.598$\pm$0.034 & 0.637$\pm$0.094 \\
DD & \textbf{0.788$\pm$0.006} & 0.703$\pm$0.034 & 0.753$\pm$0.043 & 0.712$\pm$0.024 & 0.704$\pm$0.011 & 0.711$\pm$0.022 & 0.766$\pm$0.016 \\
MUTAG & \textbf{0.898$\pm$0.005} & 0.732$\pm$0.021 & 0.833$\pm$0.023 & 0.766$\pm$0.018 & 0.711$\pm$0.002 & 0.657$\pm$0.021 & 0.891$\pm$0.057 \\
PROTEINS & \textbf{0.794$\pm$0.026} & 0.699$\pm$0.056 & 0.761$\pm$0.034 & 0.733$\pm$0.034 & 0.621$\pm$0.025 & 0.656$\pm$0.024 & 0.736$\pm$0.029 \\
NCI1 & 0.701$\pm$0.024 & 0.683$\pm$0.045 & 0.653$\pm$0.028 & \textbf{0.713$\pm$0.044} & 0.504$\pm$0.018 & 0.637$\pm$0.012 & 0.592$\pm$0.006 \\
DHFR & 0.624$\pm$0.056 & 0.612$\pm$0.026 & 0.618$\pm$0.022 & \textbf{0.634$\pm$0.033} & 0.517$\pm$0.015 & 0.598$\pm$0.023 & 0.609$\pm$0.084 \\ \hline
\textbf{Avg.\ Rank} & \textbf{1.25} & 4.63 & 3.38 & 2.50 & 6.50 & 5.63 & 4.13 \\ \hline
\end{tabular}
\caption{Anomaly detection performance in terms of AUC (mean ± std).
Parameter settings for all baseline methods follow their original papers.}
\label{tab:auc_results}
\end{table*}
In terms of evaluation, we use the popular anomaly detection evaluation metric - Area Under the Receiver Operating Characteristic Curve (AUC).
Higher AUC indicates better performance. We report the mean AUC and standard deviation based on 5-fold cross-validation for all datasets.

As shown in Table~\ref{tab:auc_results}, we observe that ProtoGLAD achieves the best performance on most datasets, ranking first on 6 out of the 8 datasets.
On NCI1 and DHFR, it achieves a competitive second-place performance, ranking only slightly behind SIGNET.
Overall, ProtoGLAD achieves the best average rank (1.25), indicating the effectiveness of our method in detecting anomalies.

\subsection{Interpretability}

\begin{table}[t]
\centering
\begin{tabular}{c|c|c}
\hline
\textbf{Rank} & \textbf{Anomalous Graph} & \textbf{Normal Prototype} \\
\hline
1 
& \includegraphics[width=0.12\textwidth]{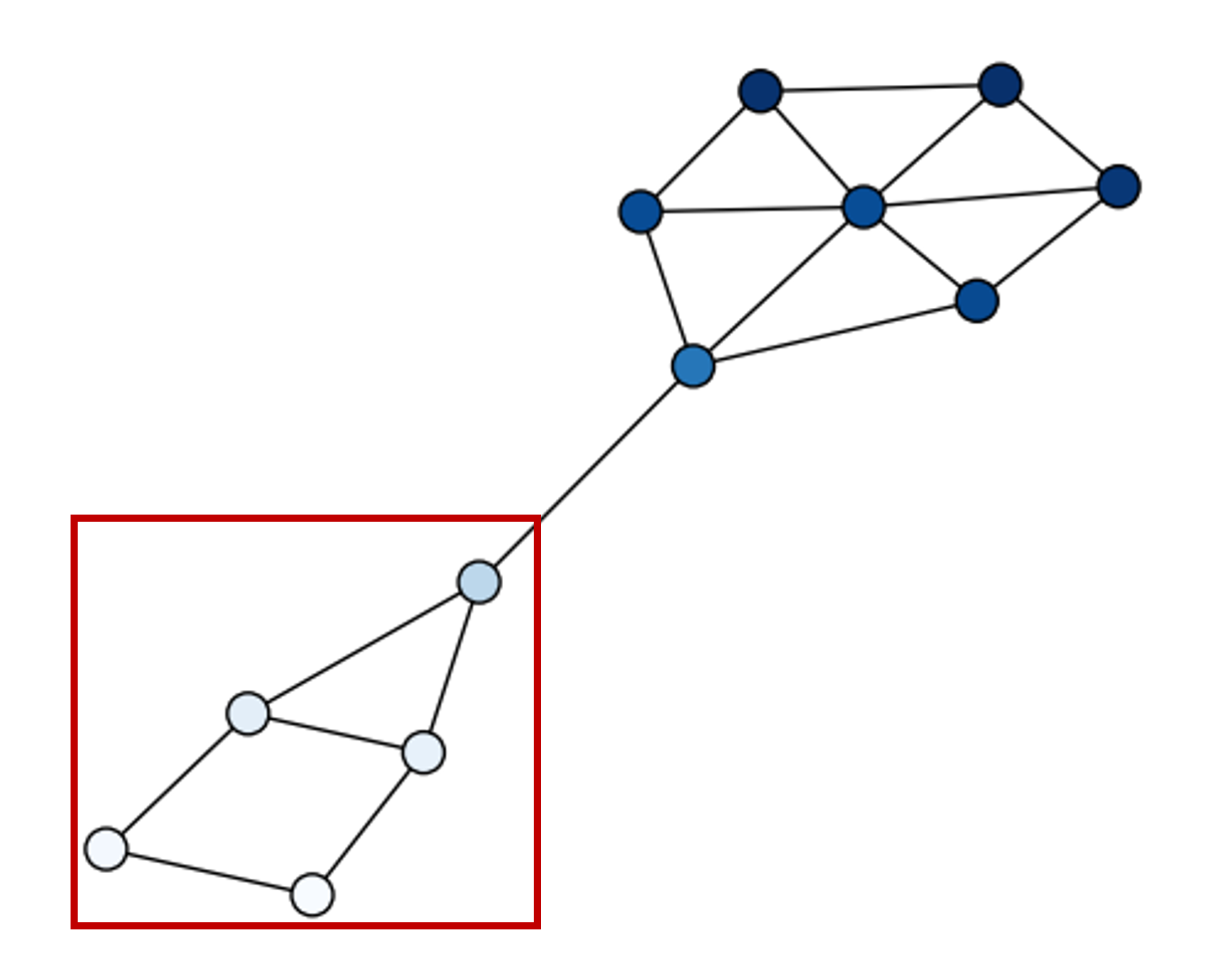} 
& \includegraphics[width=0.12\textwidth]{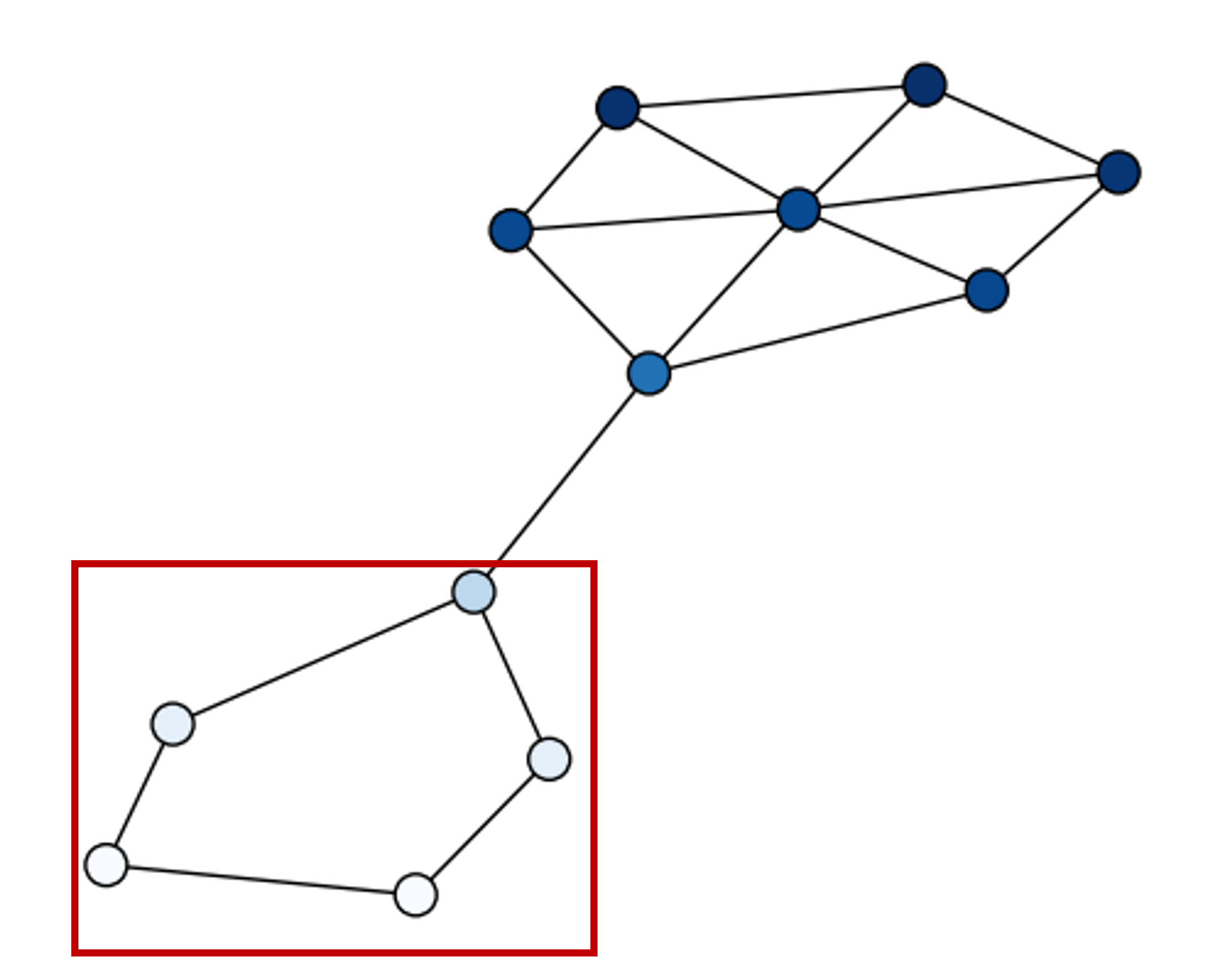} \\
\hline
2
& \includegraphics[width=0.12\textwidth]{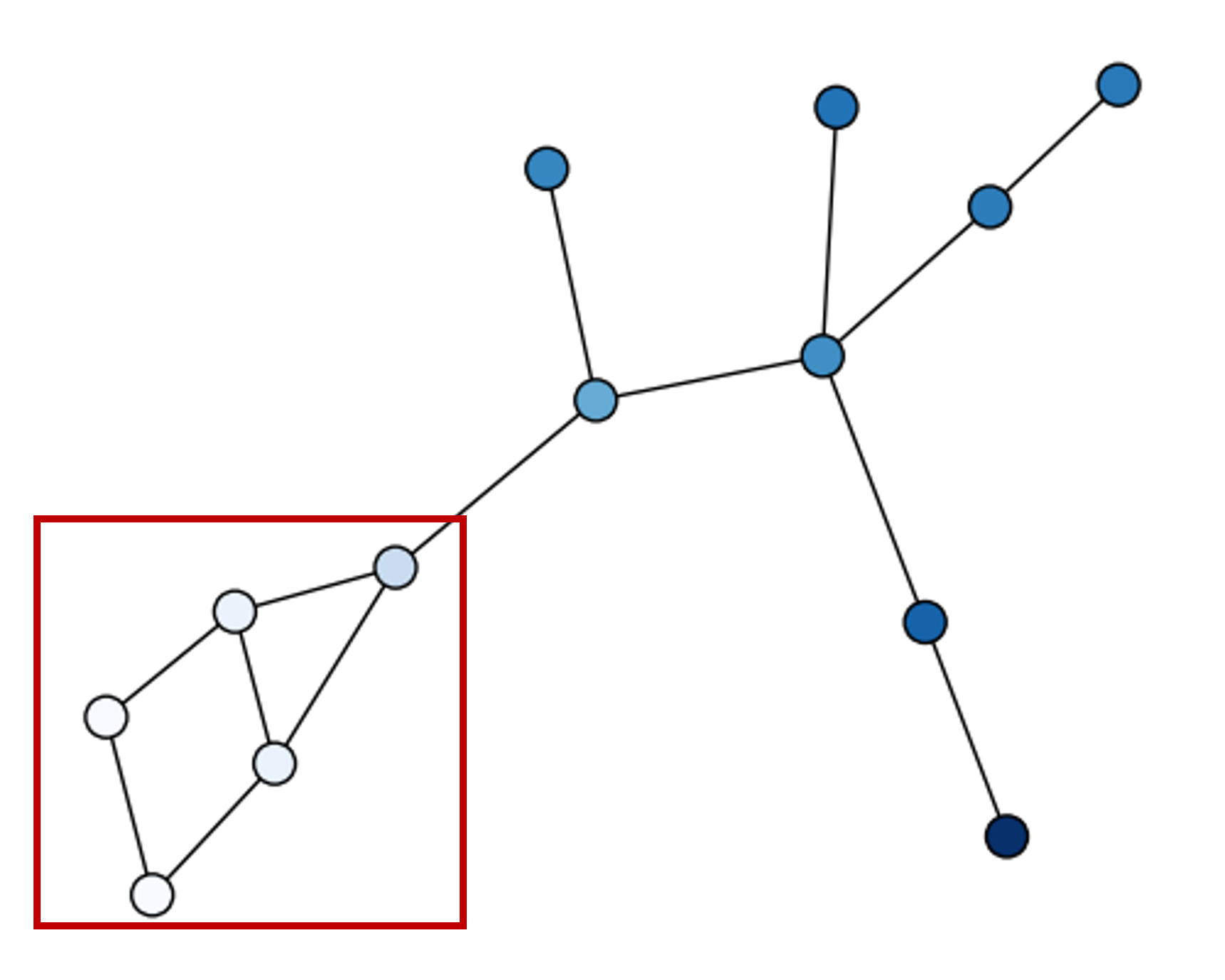} 
& \includegraphics[width=0.12\textwidth]{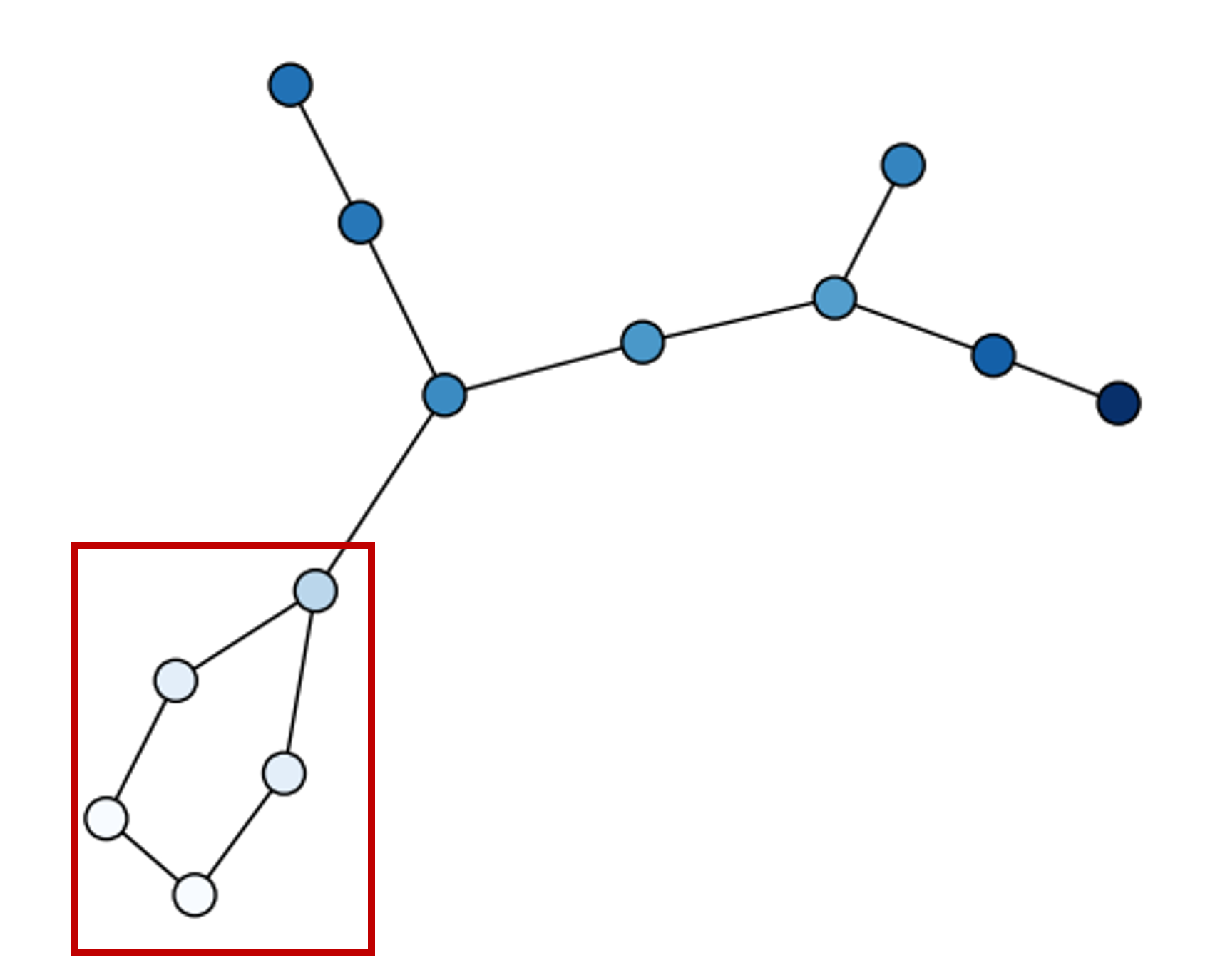} \\
\hline
3
& \includegraphics[width=0.12\textwidth]{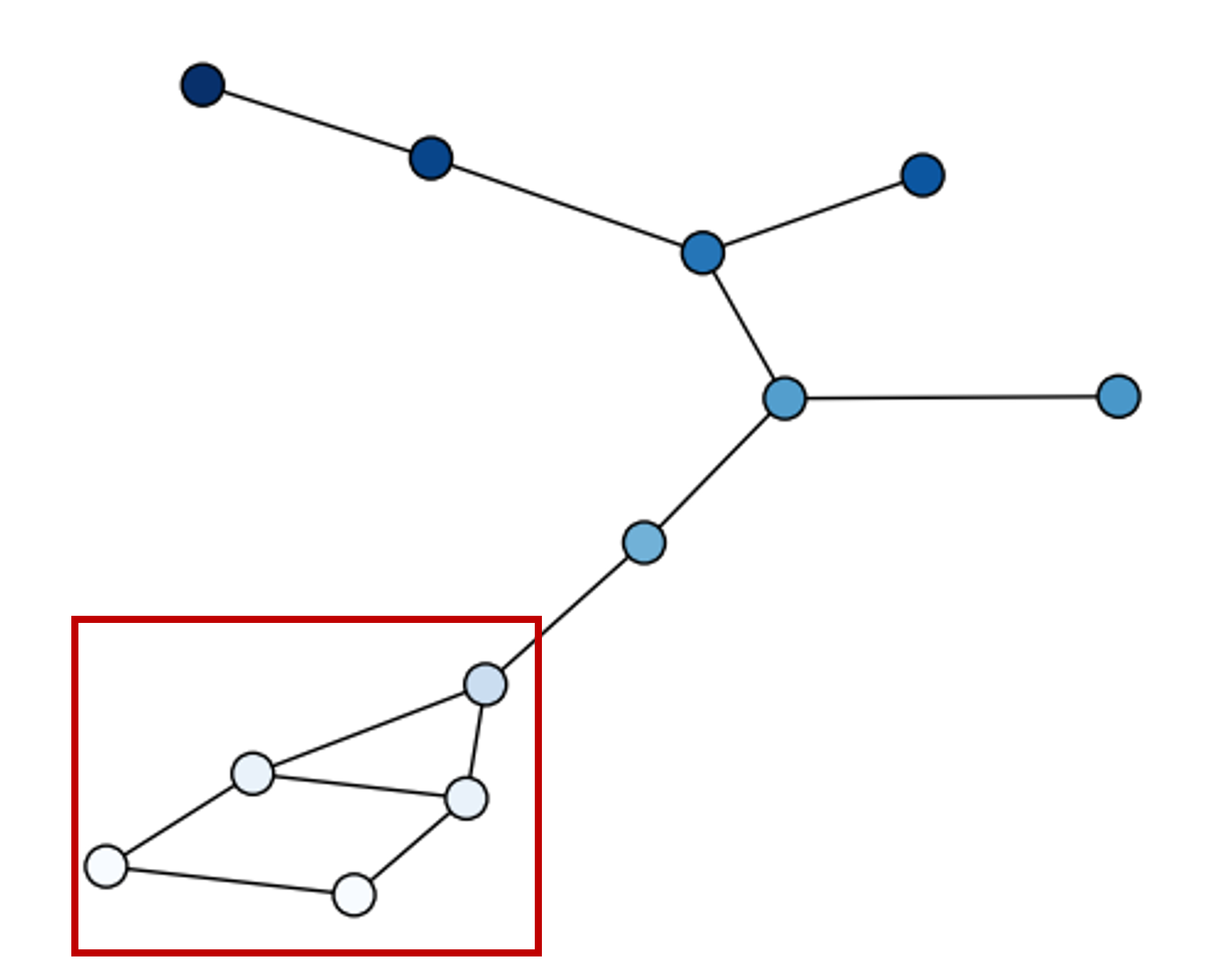} 
& \includegraphics[width=0.12\textwidth]{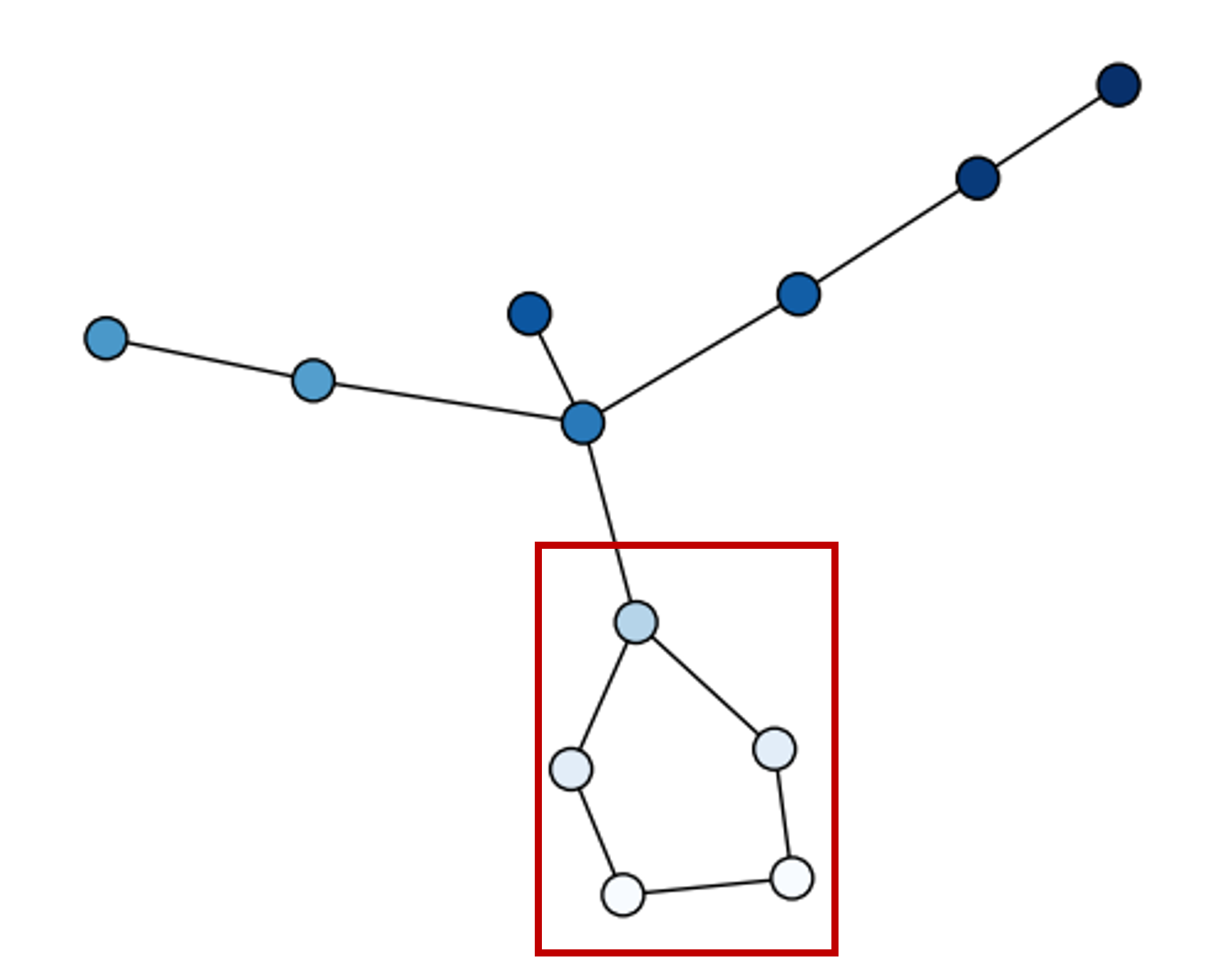} \\
\hline
\end{tabular}
\caption{Visualization of anomalous graphs and their most similar normal prototypes from the artificial dataset detected by \textbf{ProtoGLAD}.
Darker node colors indicate higher similarity between matched nodes, while lighter colors highlight potential anomaly-inducing substructures.}
\label{tab:art} 
\end{table}

To validate interpretability of ProtoGLAD, we first create a synthetic dataset inspired by previous work \cite{wu2022discovering}, where each normal graph is constructed by combining a base structure (\textit{e.g.}, tree, wheel, or ladder) with a normal substructure (circle), while anomalous graphs incorporate an anomalous substructure (house).
As shown in Table~\ref{tab:art}, we can clearly observe that our method successfully identifies matching substructures (highlighted by darker node colors) between anomalous graphs and their most similar normal prototypes, and distinctly detects anomaly-inducing substructures (highlighted by lighter node colors).
\begin{table}[!t]
\centering
\begin{tabular}{c|c|c}
\hline
\textbf{Dataset} & \textbf{Anomalous Graph} & \textbf{Normal Prototype} \\
\hline
\multirow{1}{*}{AIDS} 
& \includegraphics[width=0.12\textwidth]{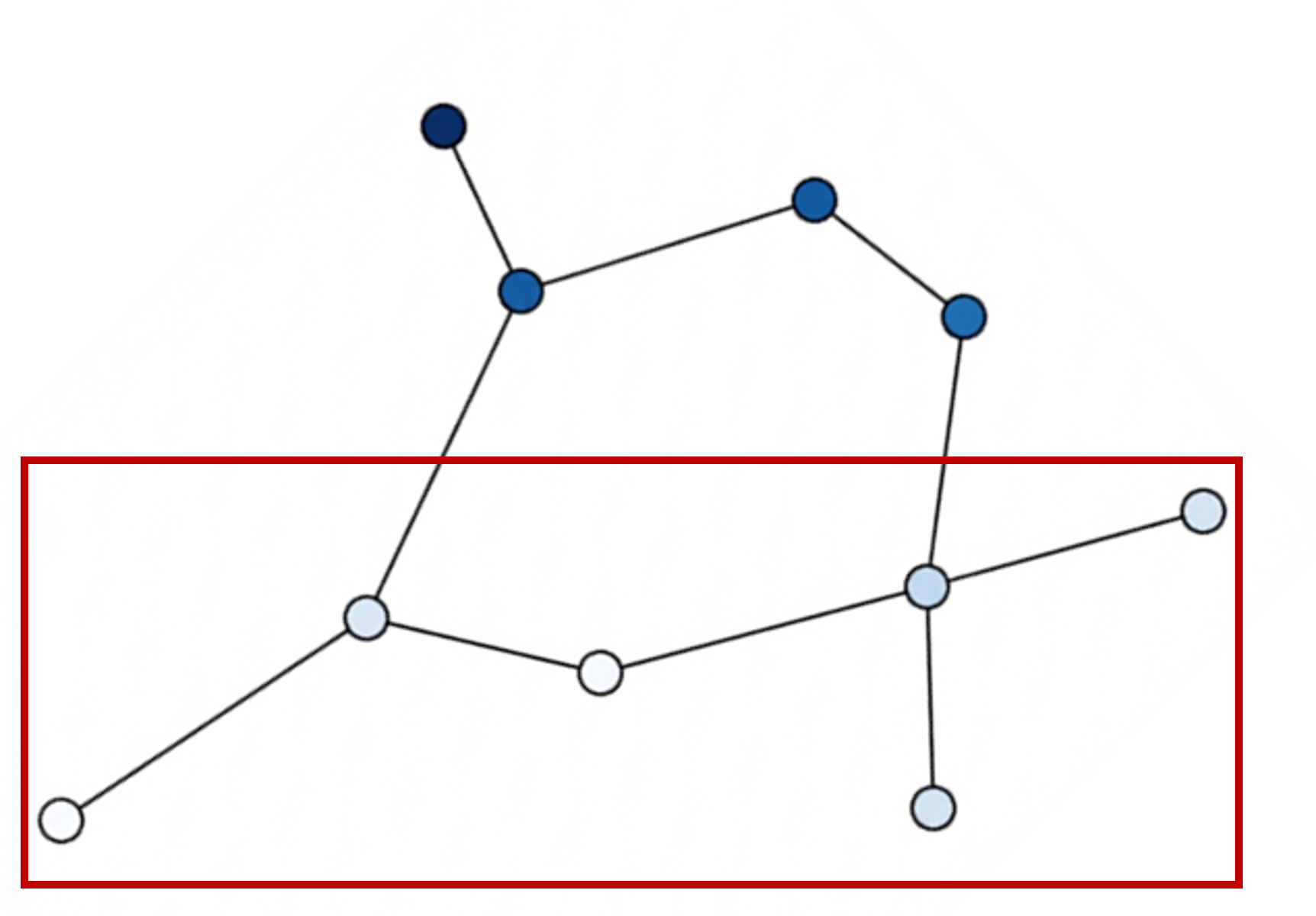} 
& \includegraphics[width=0.12\textwidth]{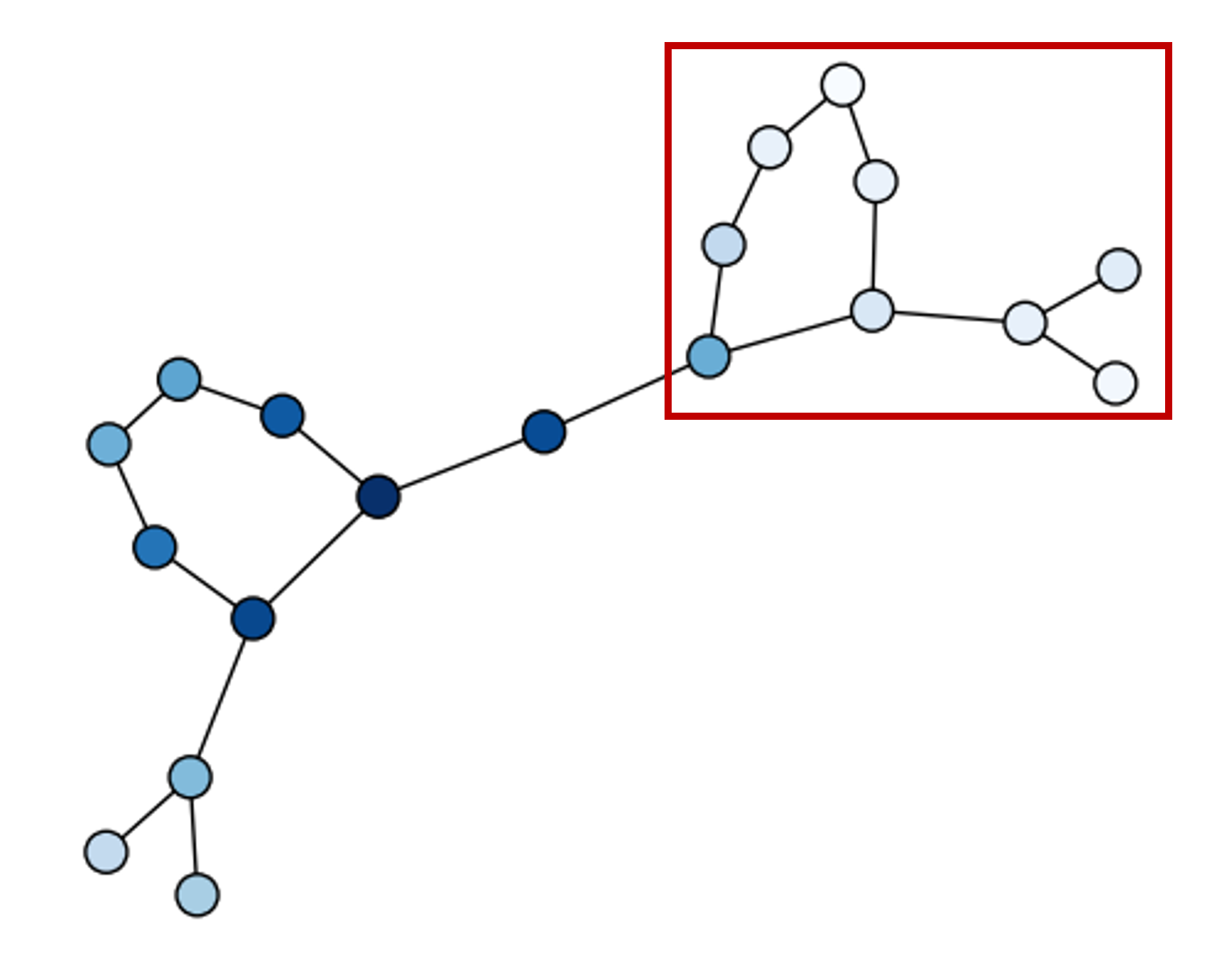}  \\
& \includegraphics[width=0.12\textwidth]{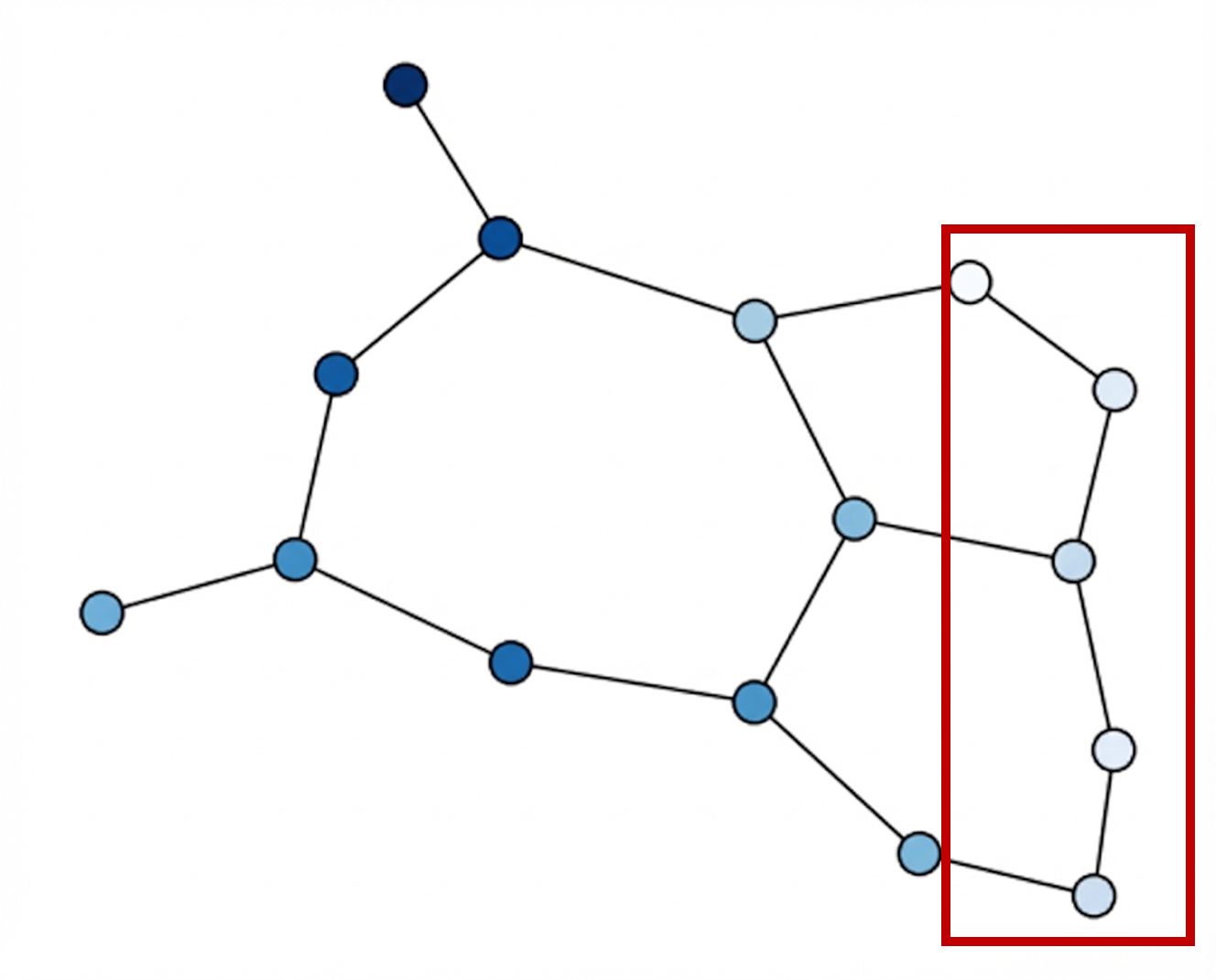} 
& \includegraphics[width=0.12\textwidth]{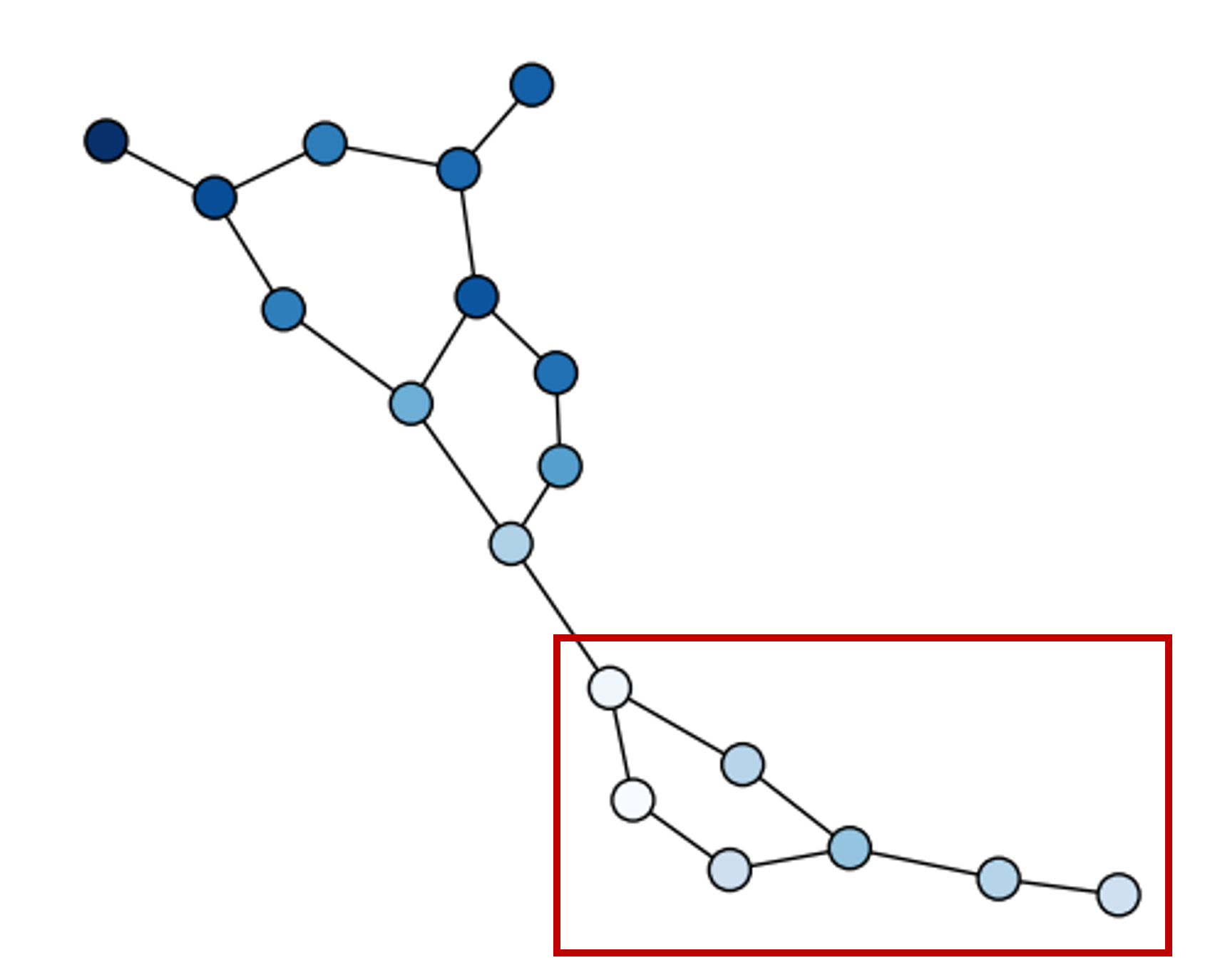}  \\
\hline
\multirow{1}{*}{COX2} 
& \includegraphics[width=0.12\textwidth]{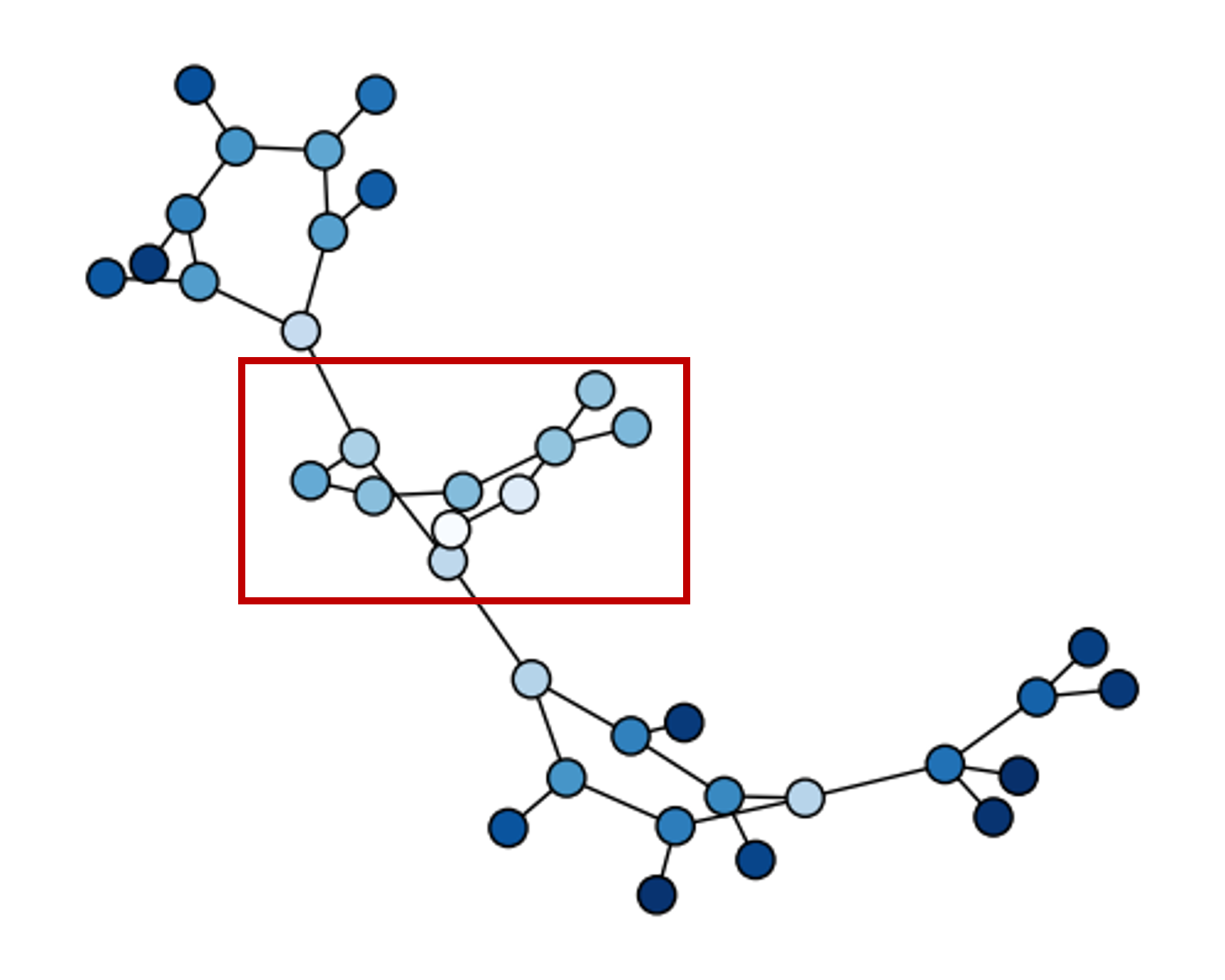} 
& \includegraphics[width=0.12\textwidth]{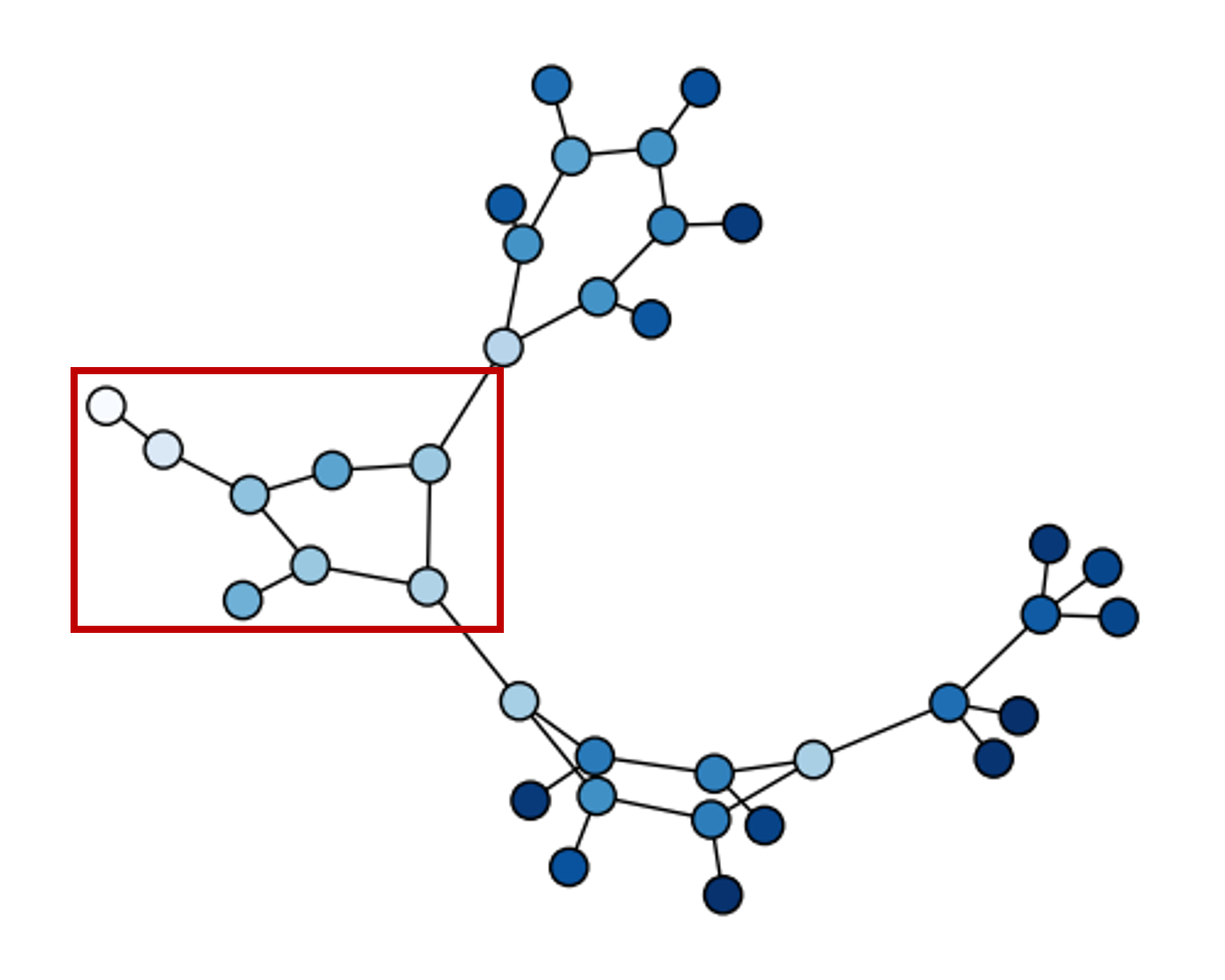}  \\
& \includegraphics[width=0.12\textwidth]{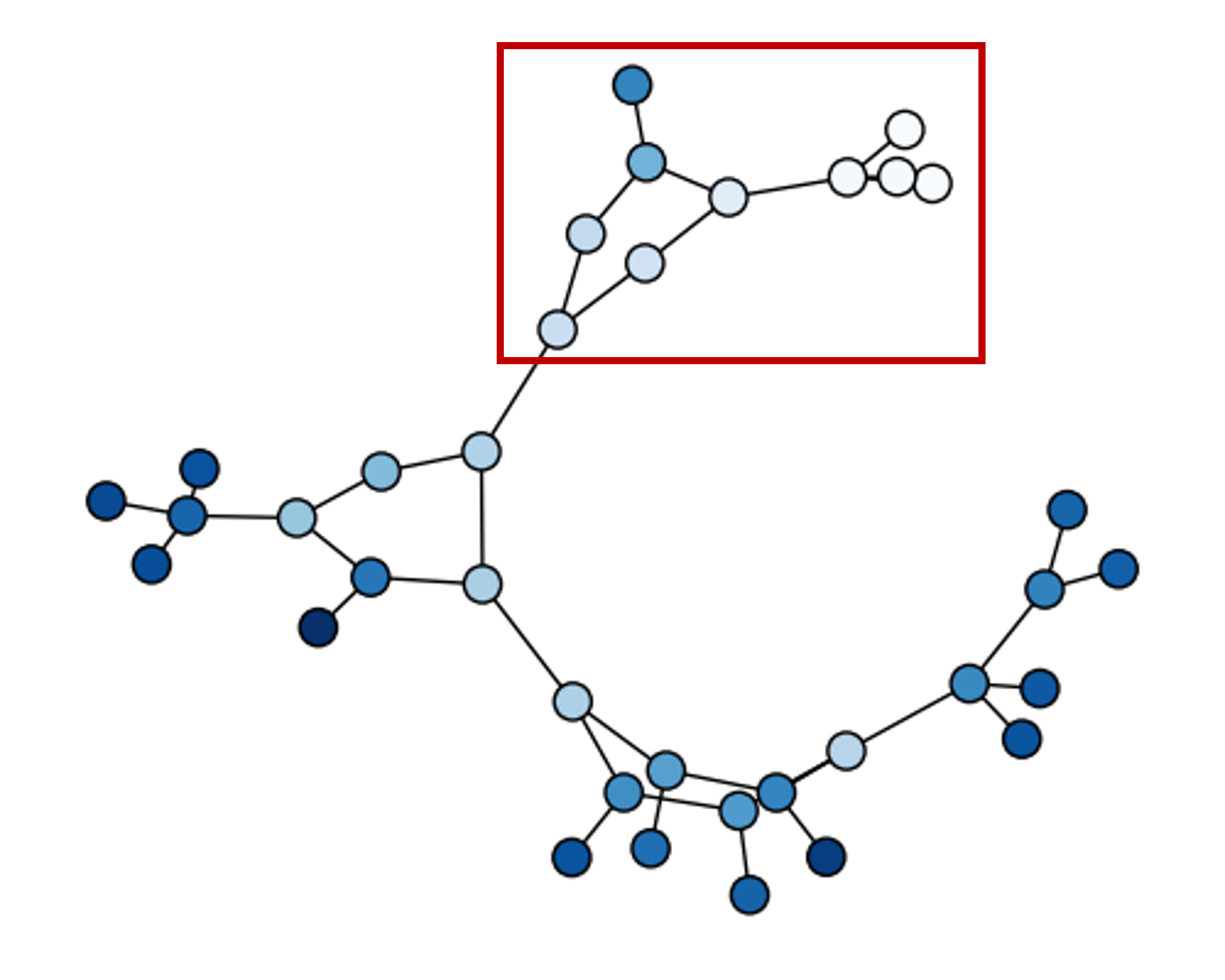} 
& \includegraphics[width=0.12\textwidth]{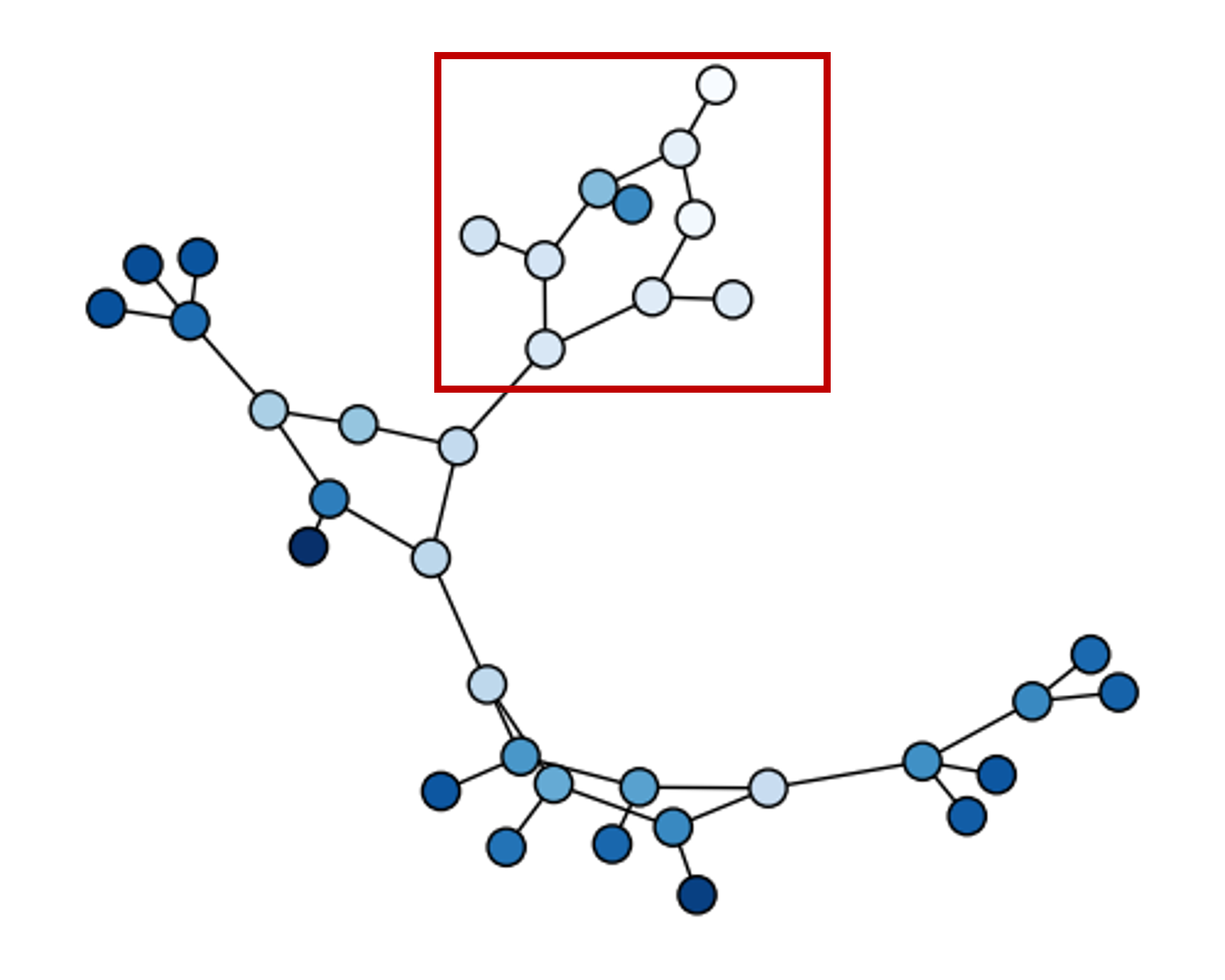}  \\
\hline
\multirow{1}{*}{PROTEINS} 
& \includegraphics[width=0.12\textwidth]{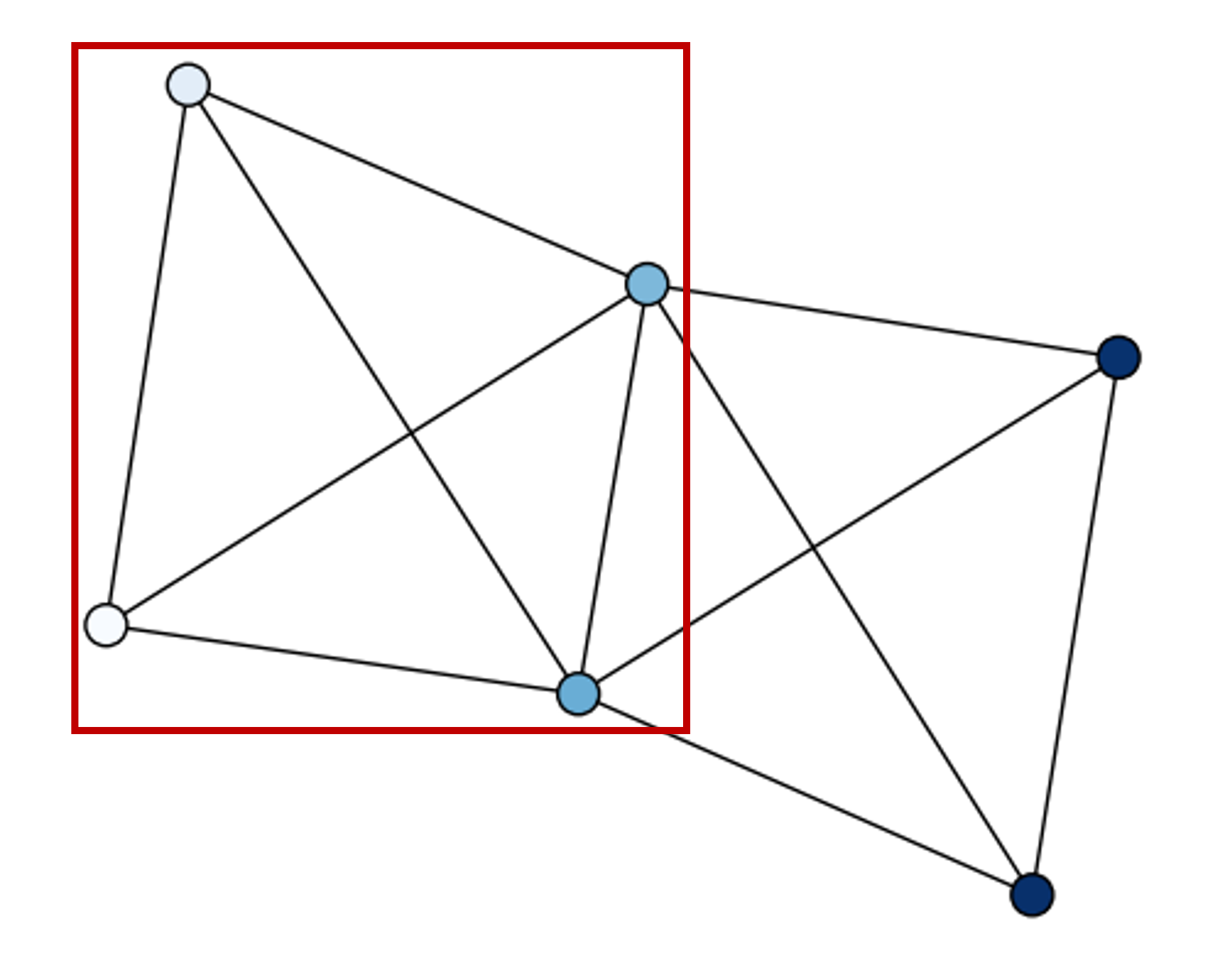} 
& \includegraphics[width=0.12\textwidth]{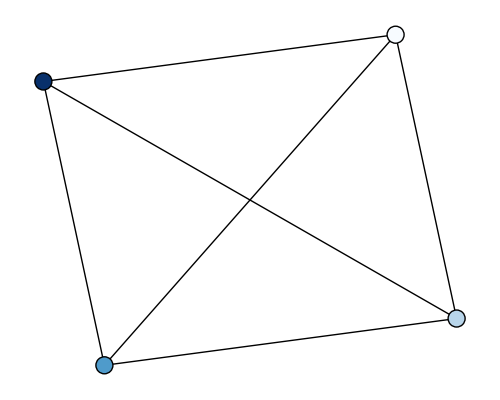} \\
& \includegraphics[width=0.12\textwidth]{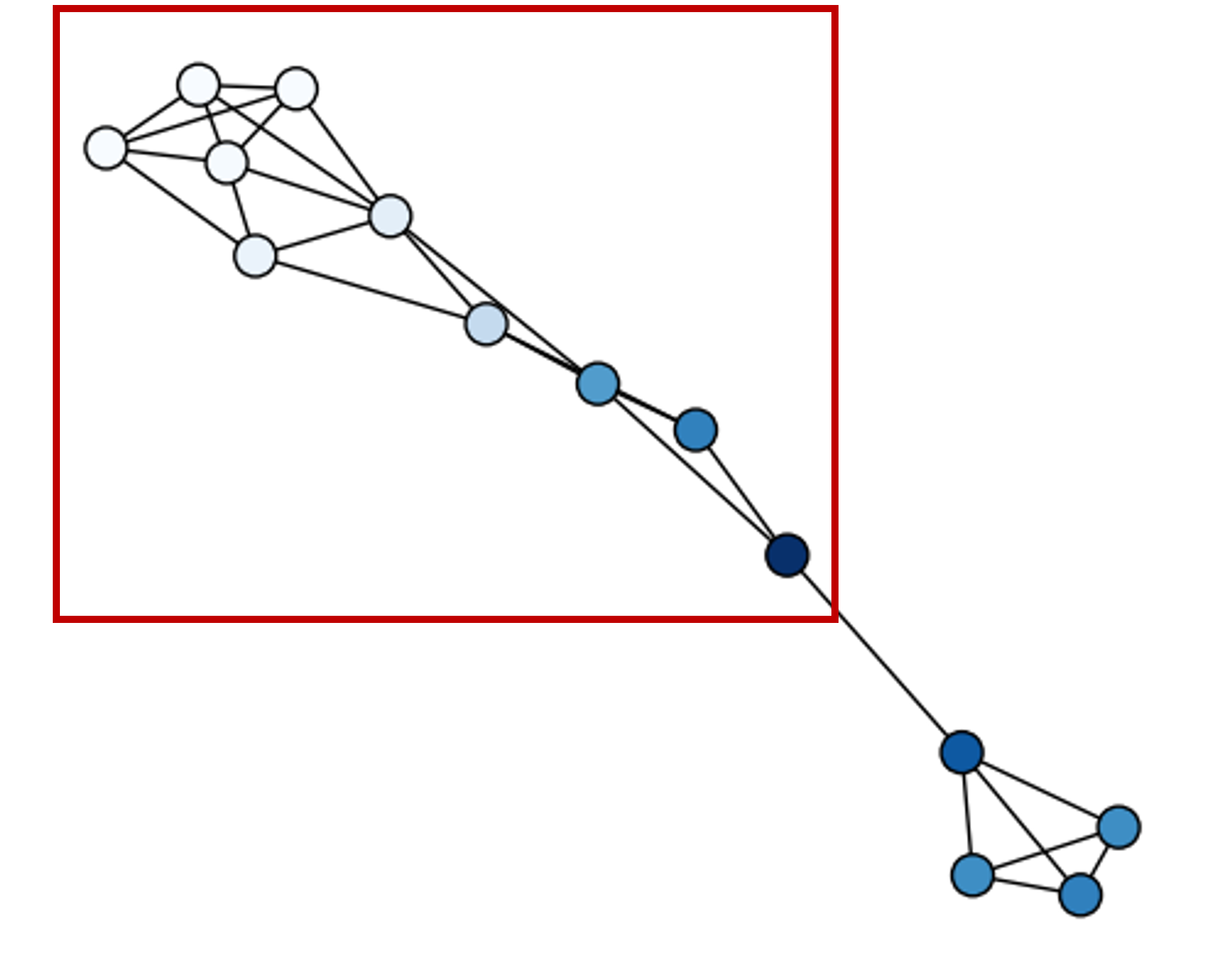} 
& \includegraphics[width=0.12\textwidth]{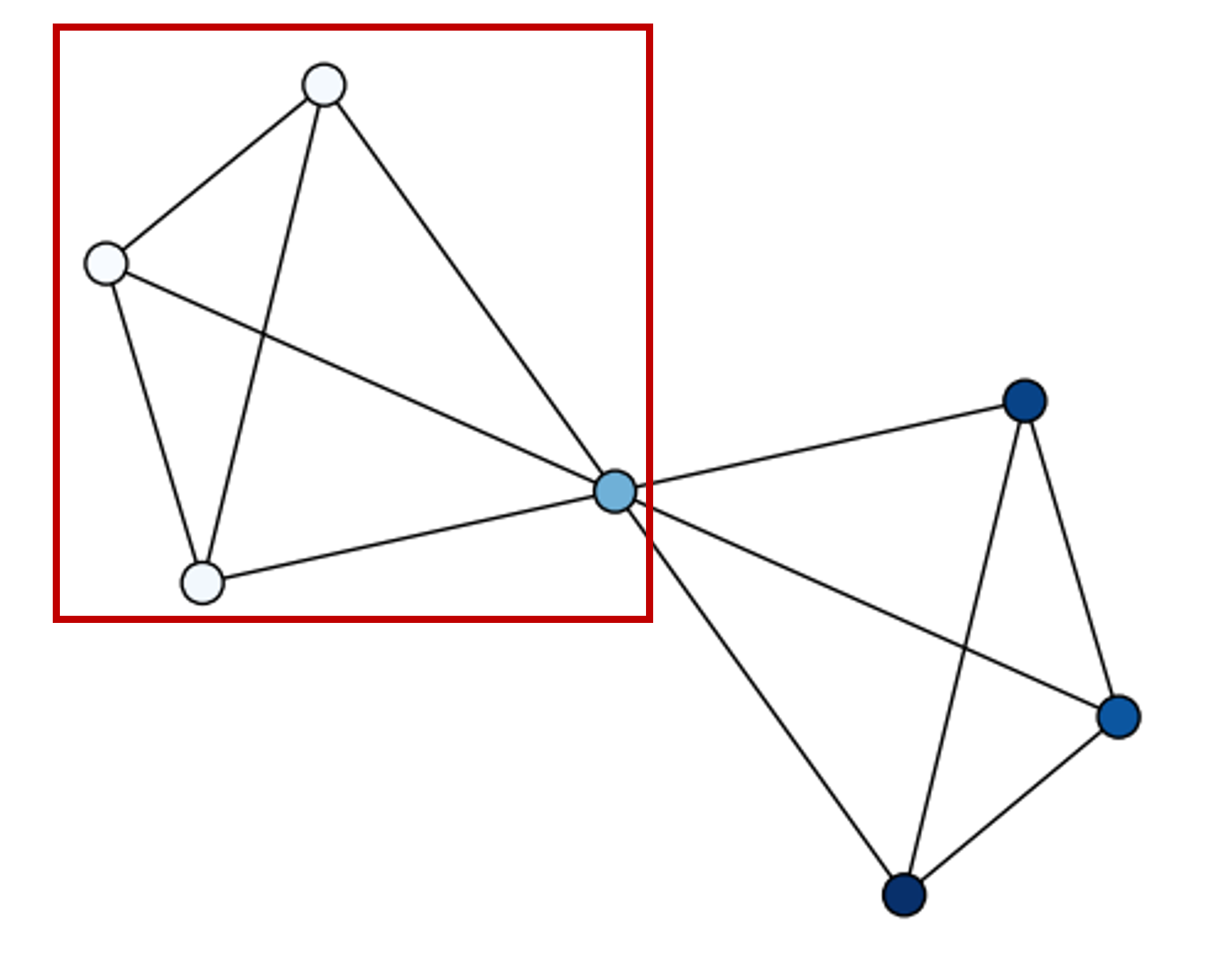} \\
\hline
\end{tabular}
\caption{Visualization of anomalous graphs and their most similar normal prototypes detected by \textbf{ProtoGLAD} on real-world datasets.}
\label{tab:explainable}
\end{table}

\begin{table*}[!t]
\centering
\begin{tabular}{c|c|c|c|c}
\hline
\textbf{Method} & \textbf{Anomalous Graph}& \textbf{Anomaly Rank} & \textbf{Normal Prototype} & \textbf{Additional Normal Reference Graphs} \\
\hline
\multirow{1}{*}{SIGNET} 
& \includegraphics[width=0.105\textwidth]{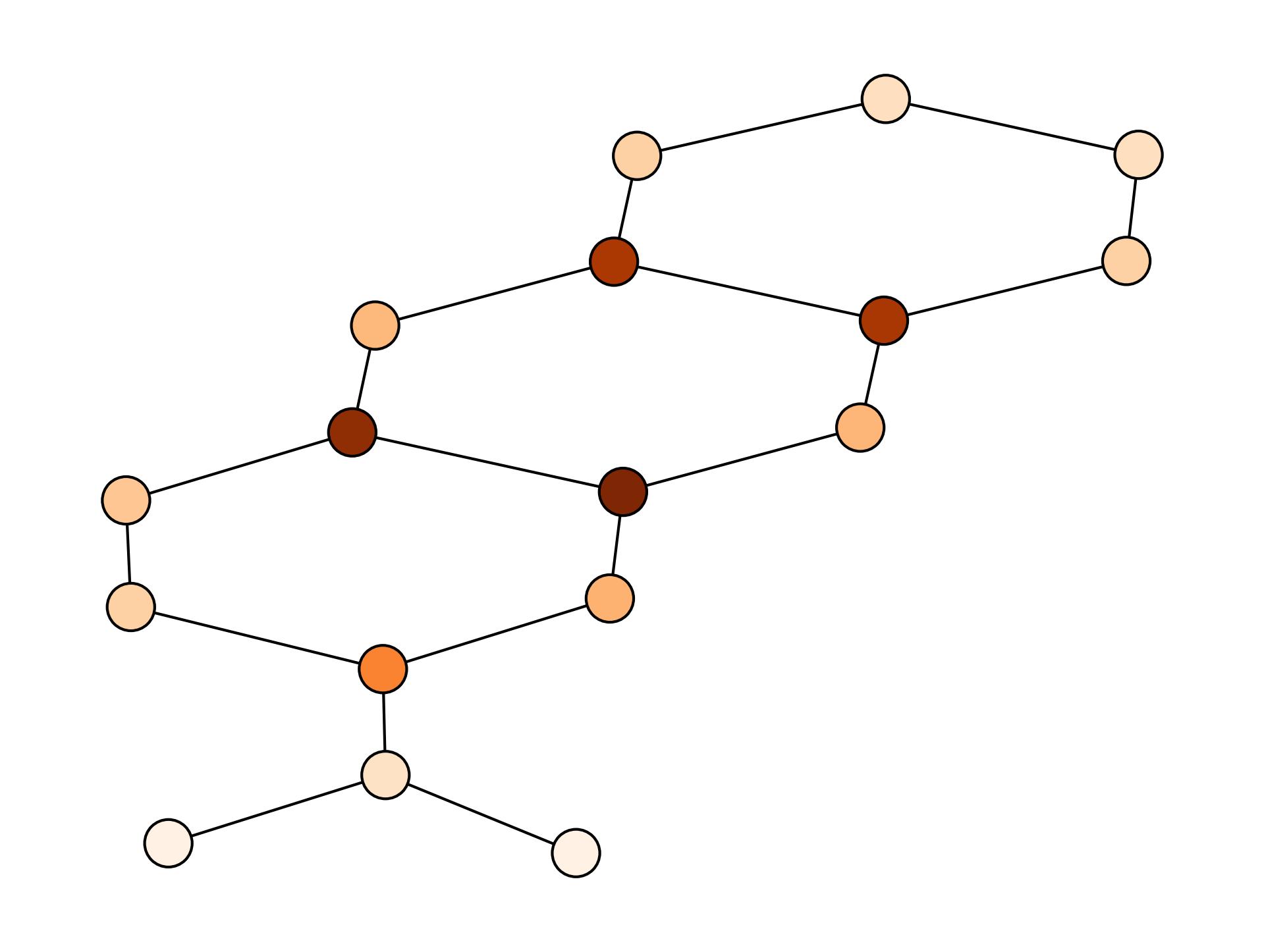} 
& \includegraphics[width=0.105\textwidth]{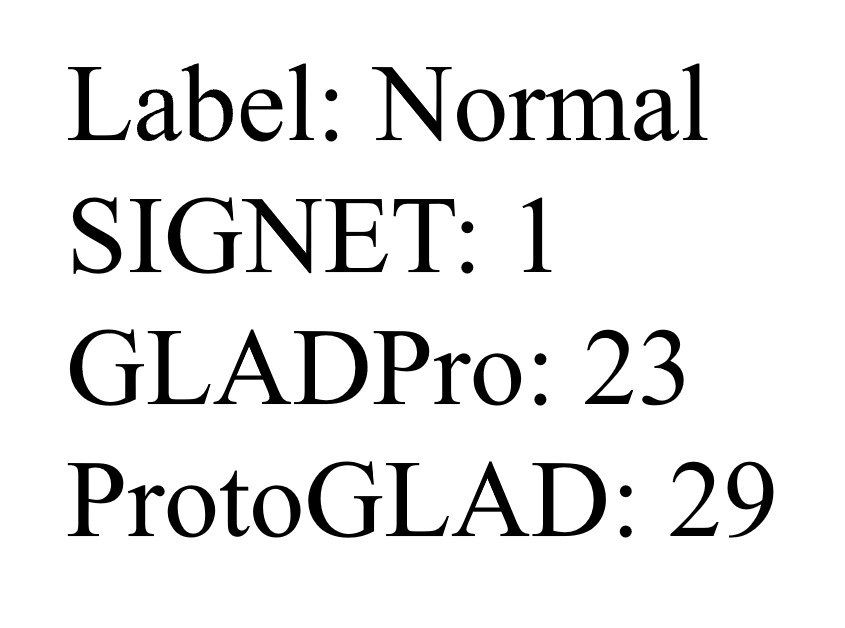} 
&  \includegraphics[width=0.115\textwidth]{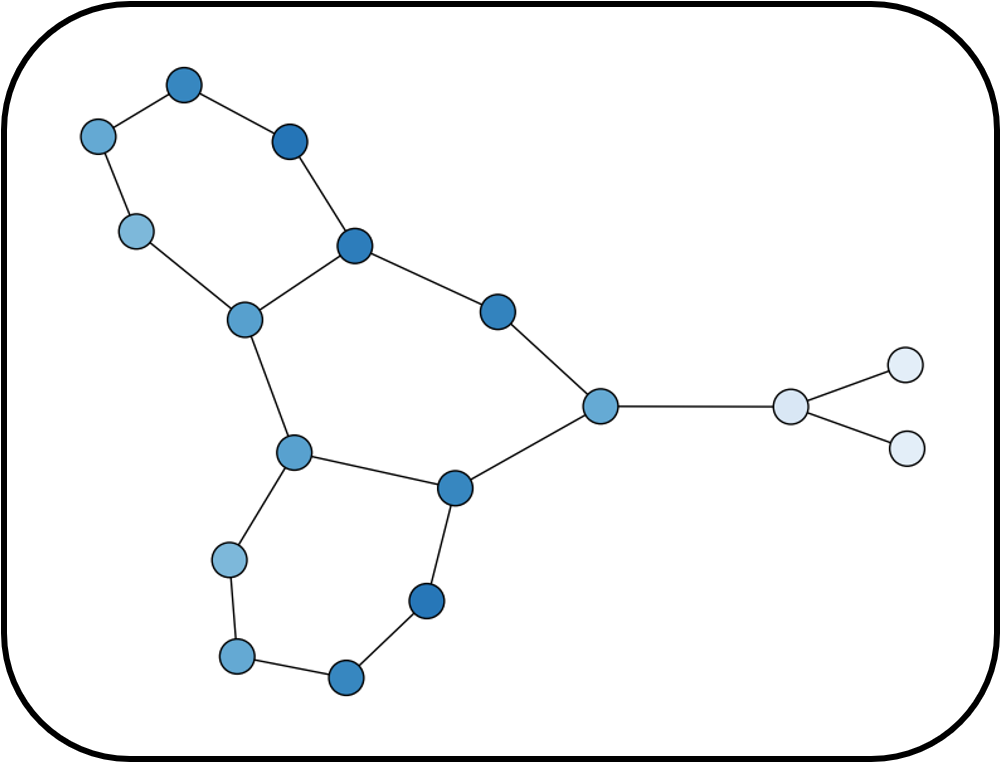} 
&  \includegraphics[width=0.36\textwidth]{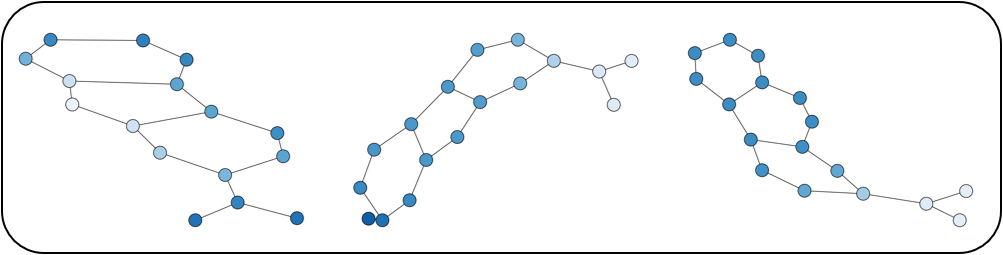}\\
& \includegraphics[width=0.105\textwidth]{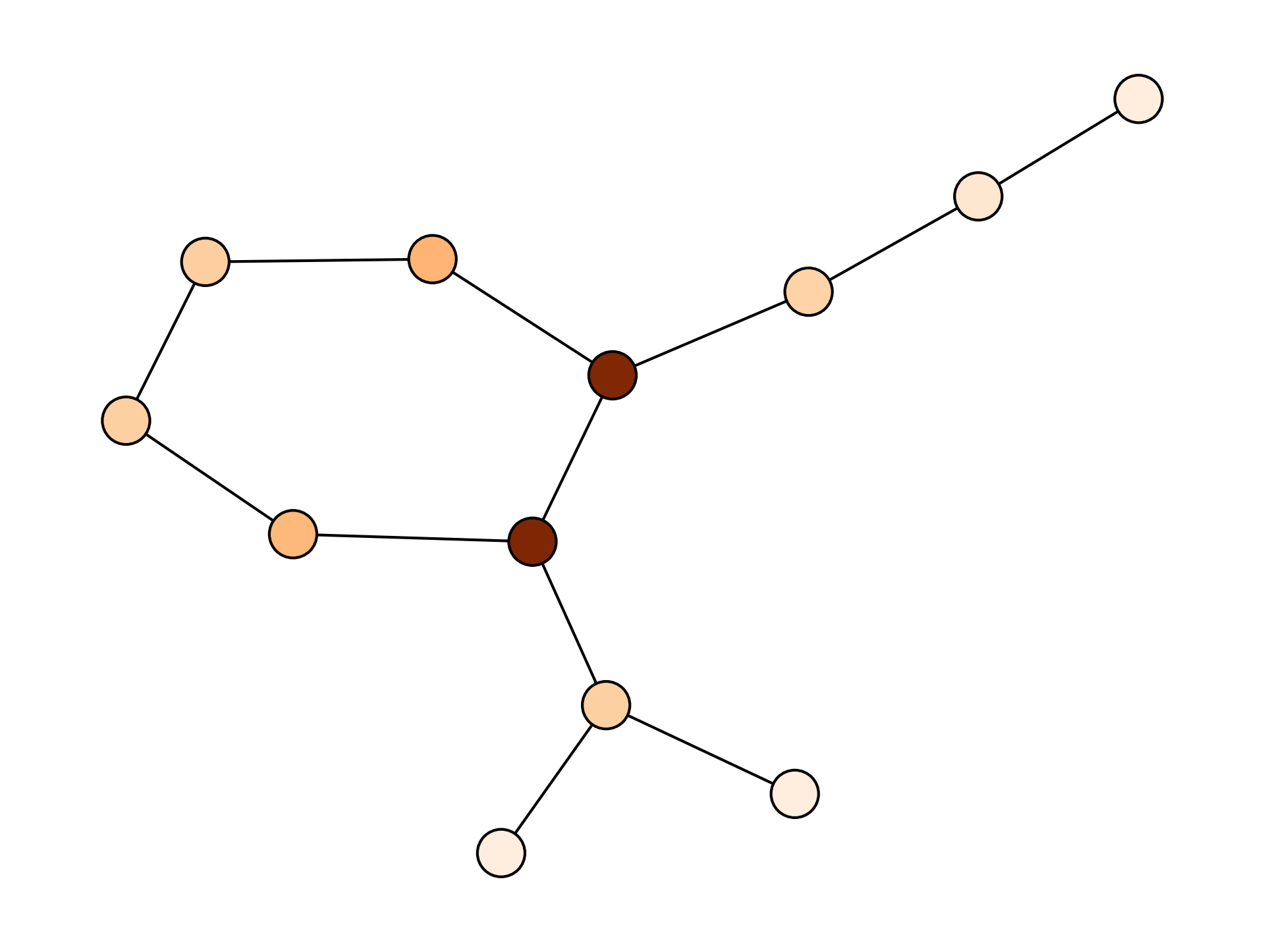} 
& \includegraphics[width=0.105\textwidth]{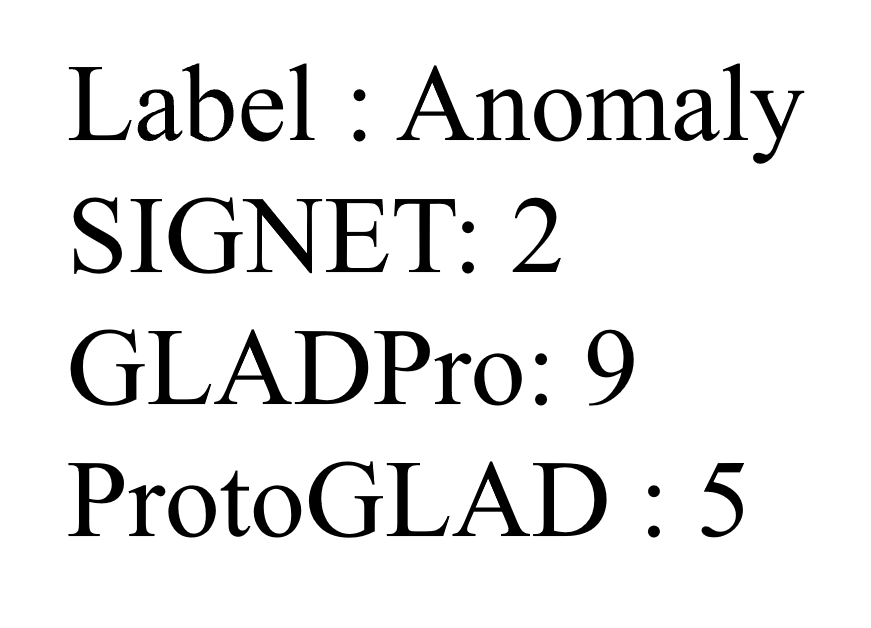} 
&  \includegraphics[width=0.115\textwidth]{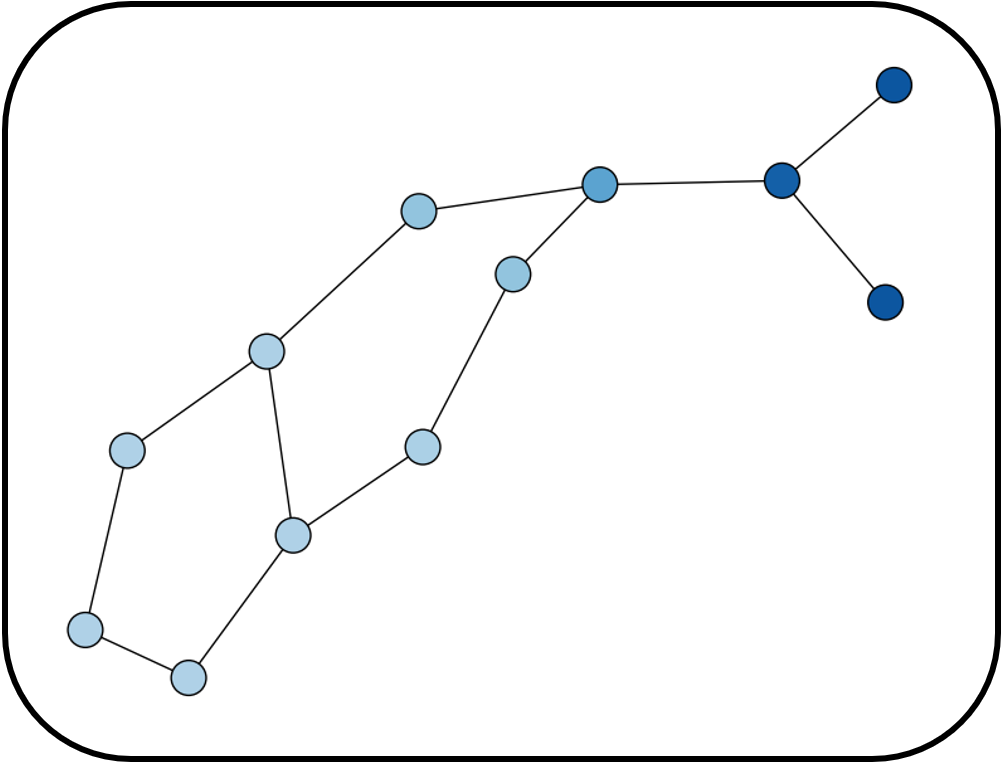} 
&  \includegraphics[width=0.36\textwidth]{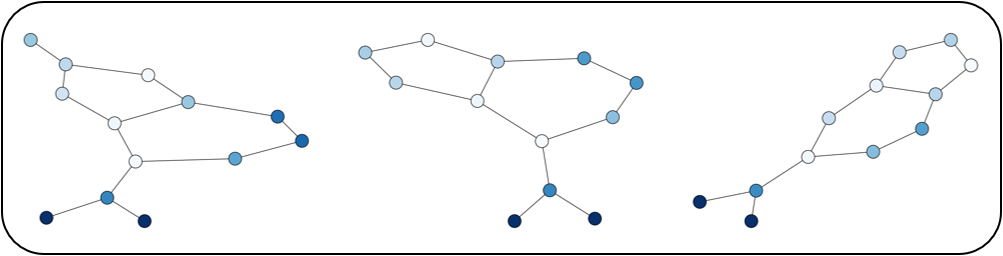}\\
\hline
\multirow{1}{*}{GLADPro} 
& \includegraphics[width=0.105\textwidth]{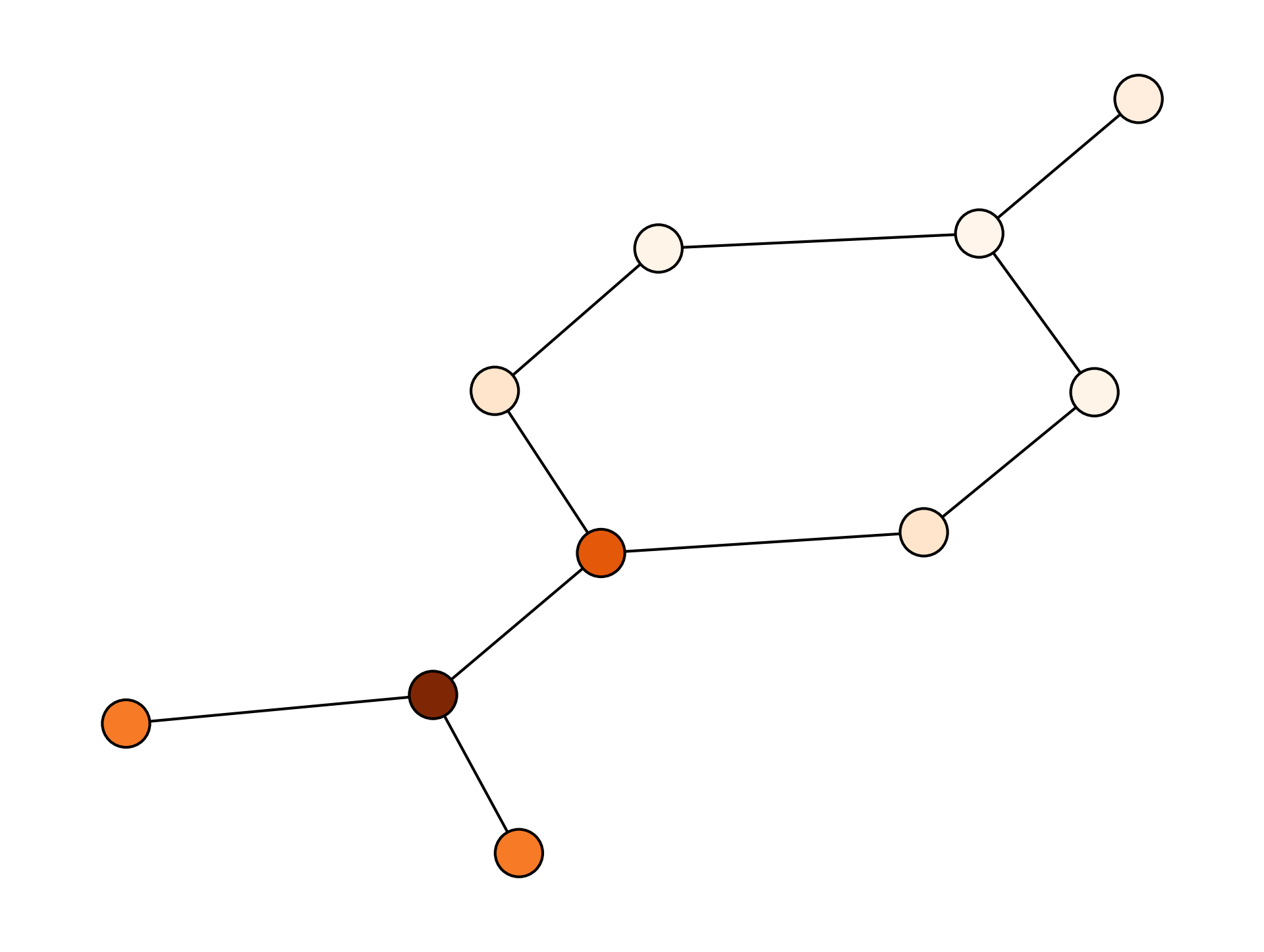} 
& \includegraphics[width=0.105\textwidth]{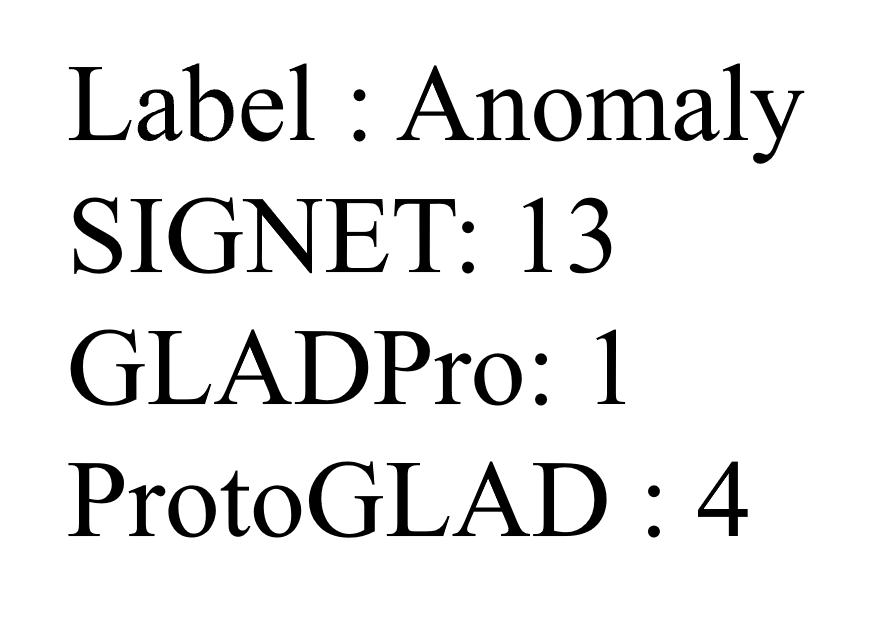}
& \includegraphics[width=0.105\textwidth]{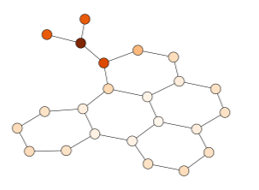}
& \includegraphics[width=0.36\textwidth]{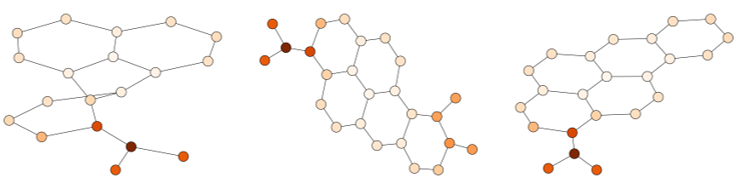} \\
& \includegraphics[width=0.105\textwidth]{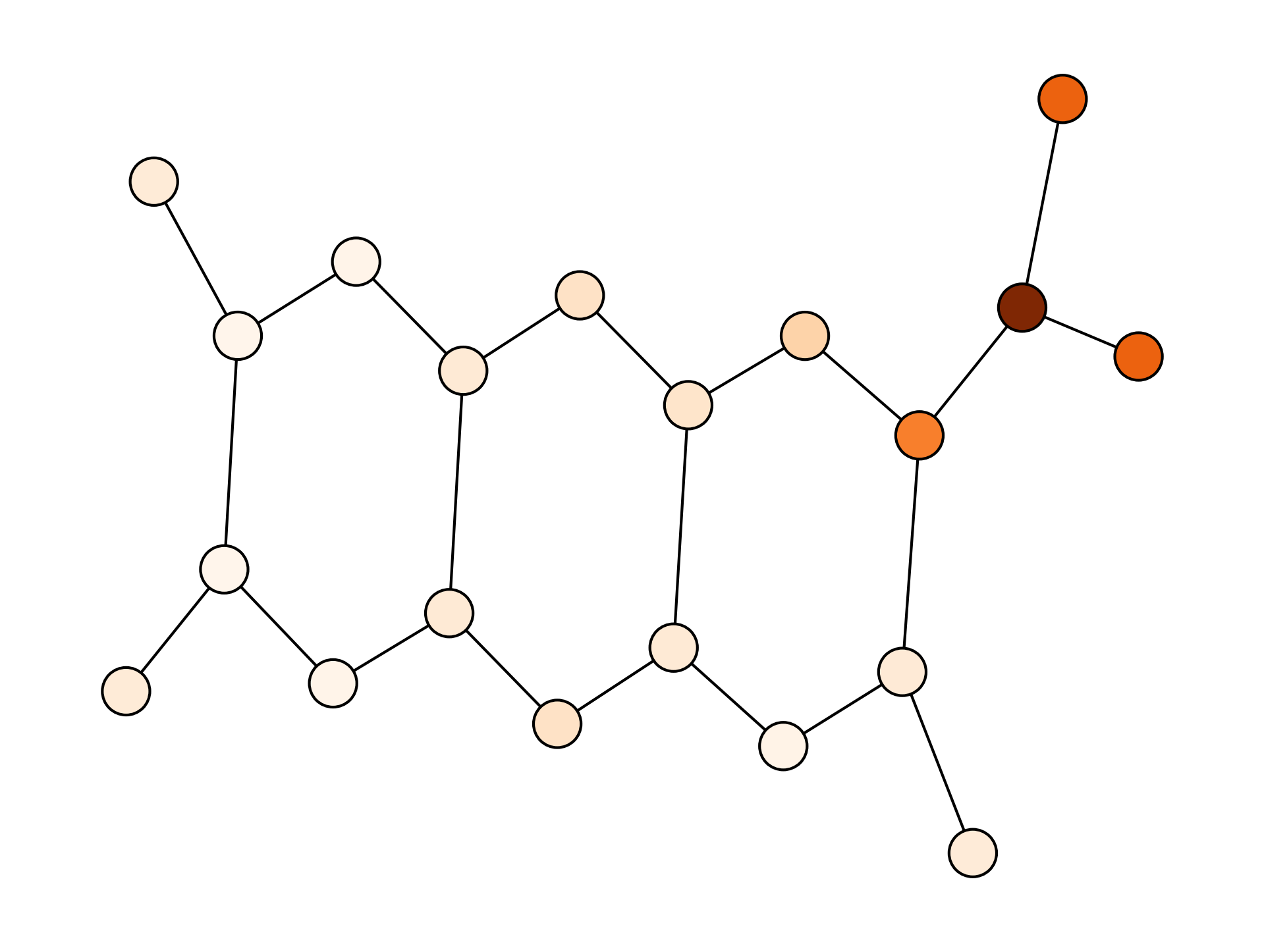} 
& \includegraphics[width=0.105\textwidth]{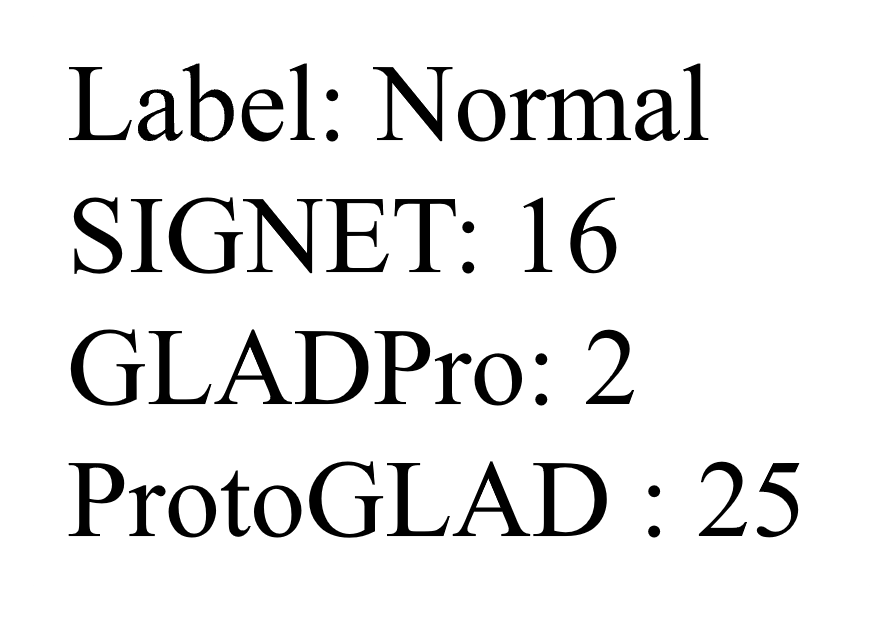}
& \includegraphics[width=0.105\textwidth]{figure/gladpro_p1.png}
& \includegraphics[width=0.36\textwidth]{figure/gladpro_p1n.png} \\
\hline
\multirow{1}{*}{ProtoGLAD} 
& \includegraphics[width=0.105\textwidth]{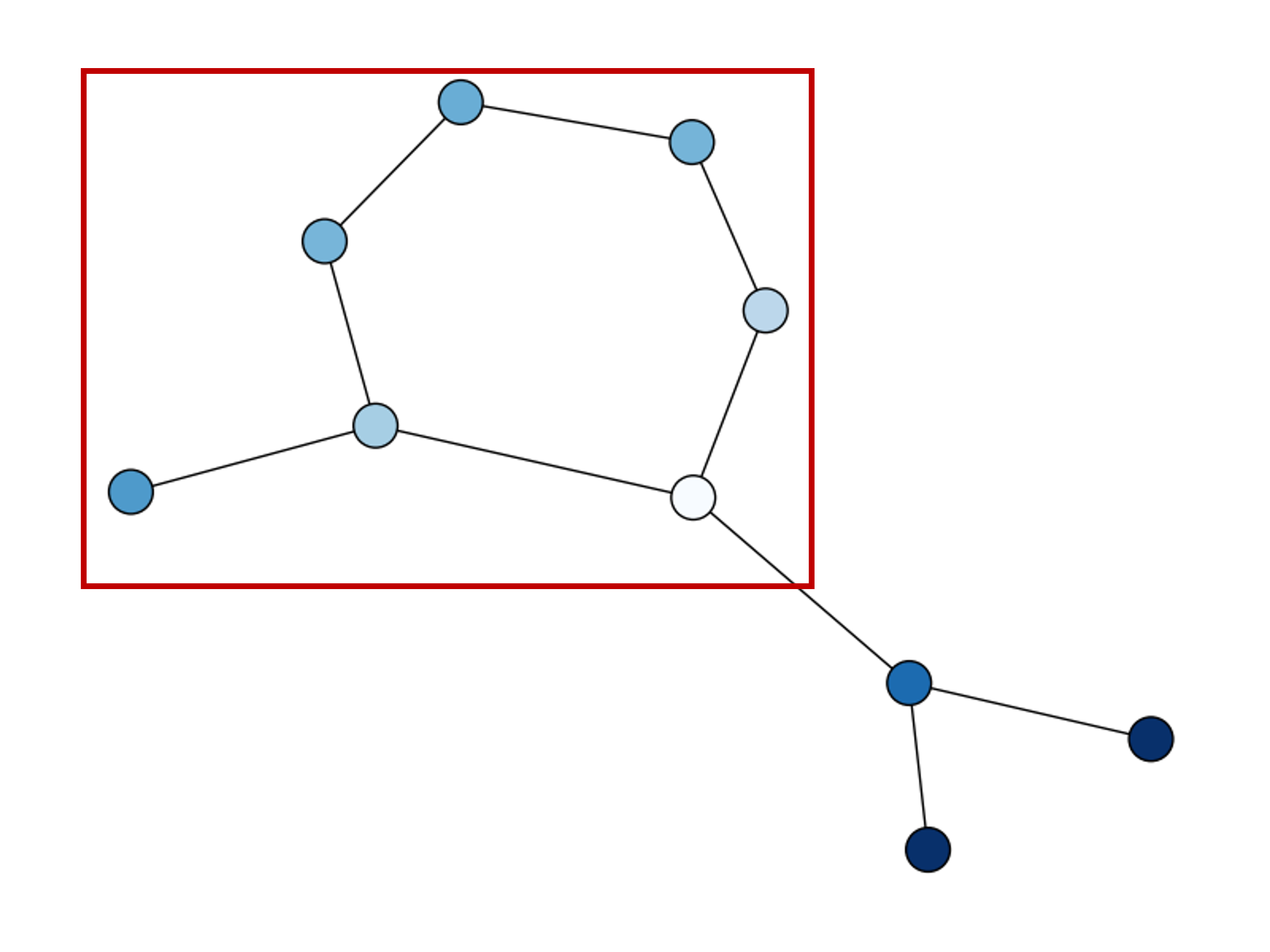} 
& \includegraphics[width=0.105\textwidth]{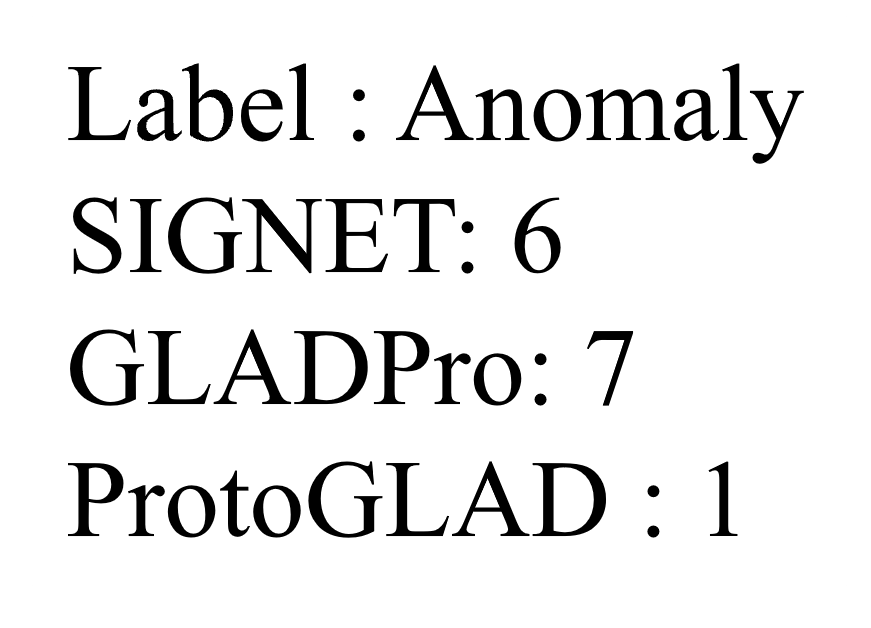}
& \includegraphics[width=0.105\textwidth]{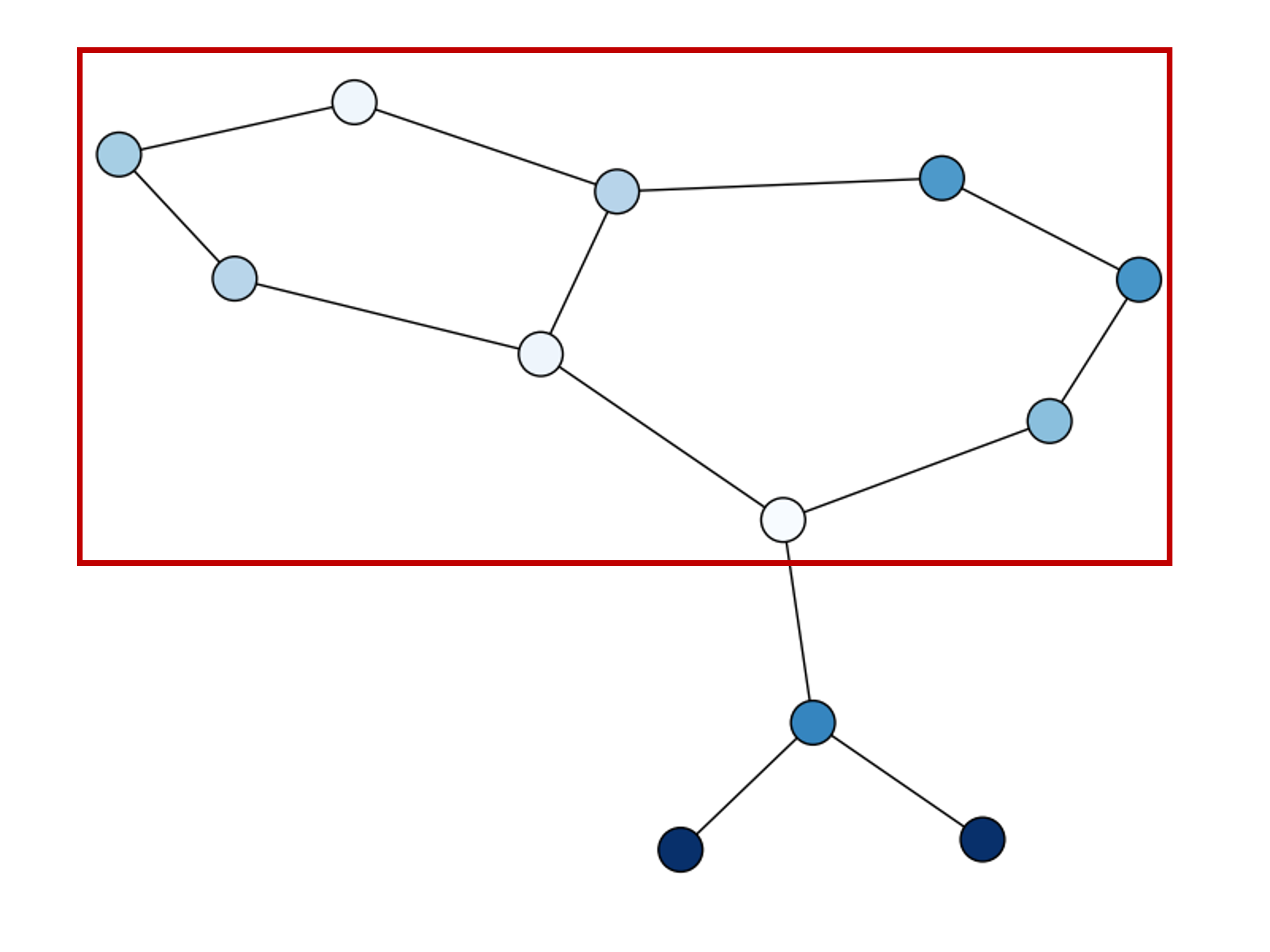} 
& \includegraphics[width=0.36\textwidth]{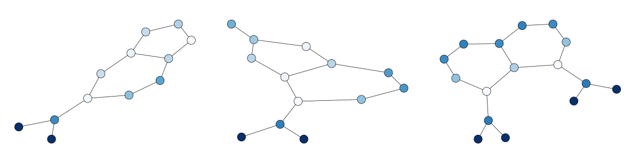} \\
& \includegraphics[width=0.105\textwidth]{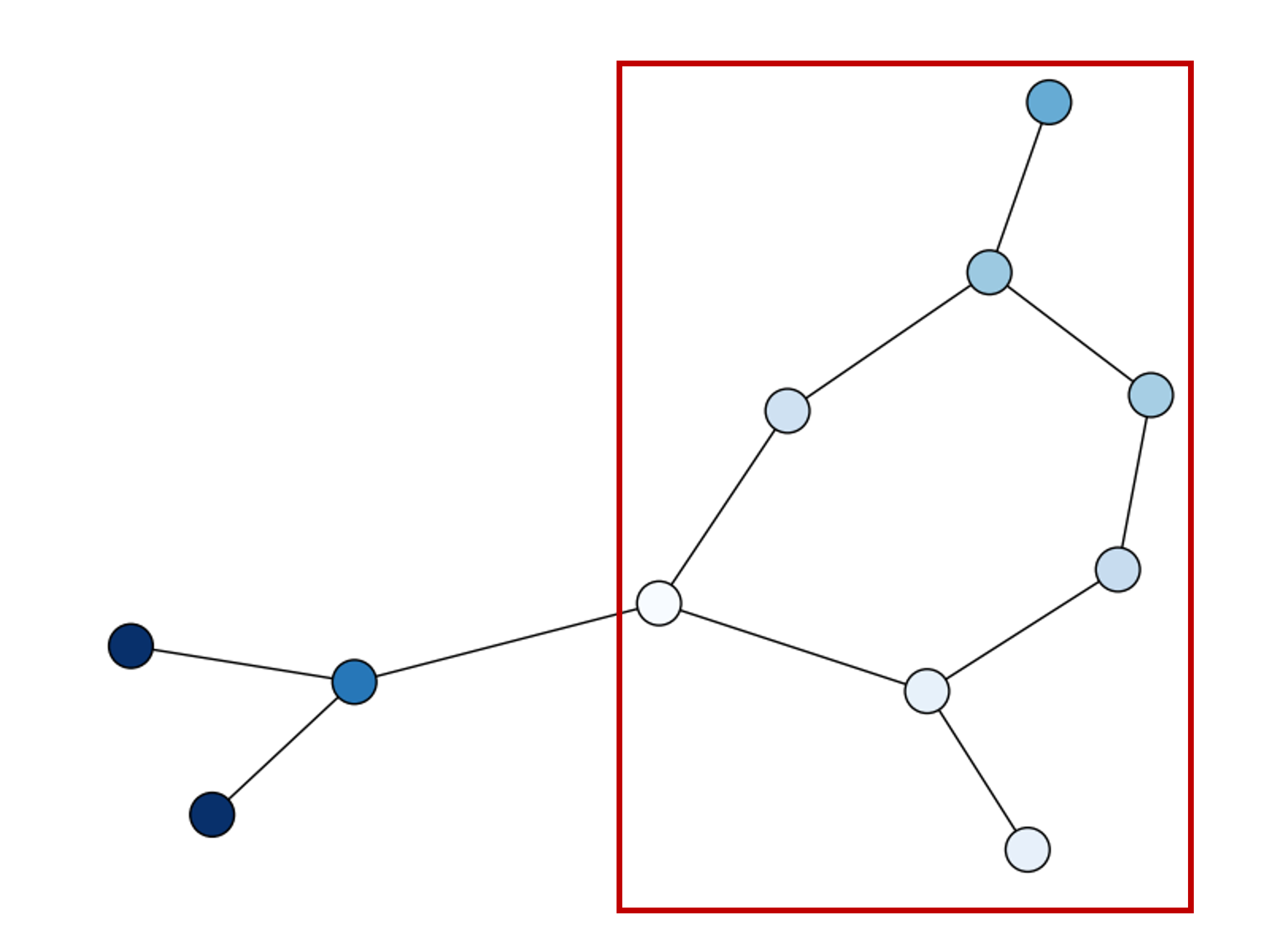} 
& \includegraphics[width=0.105\textwidth]{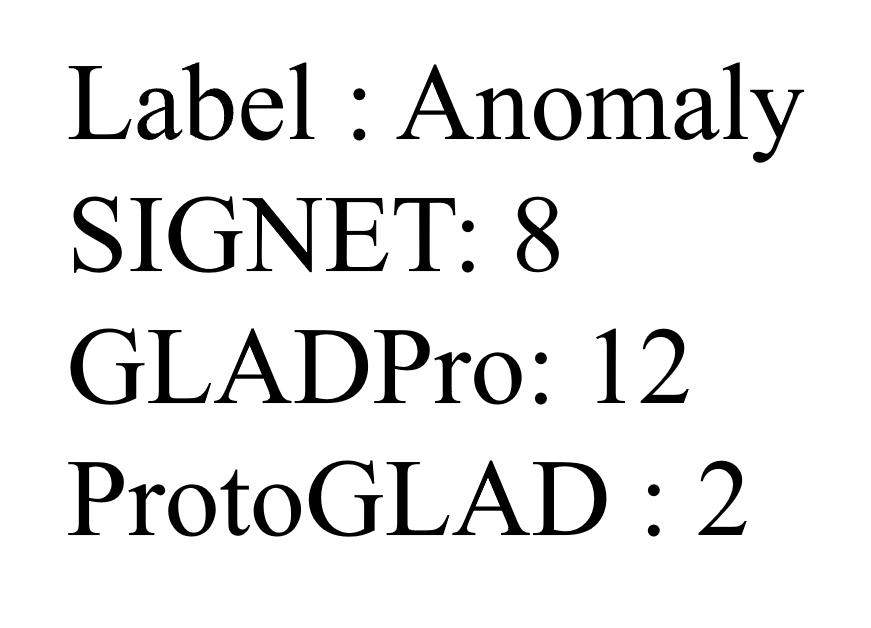}
& \includegraphics[width=0.105\textwidth]{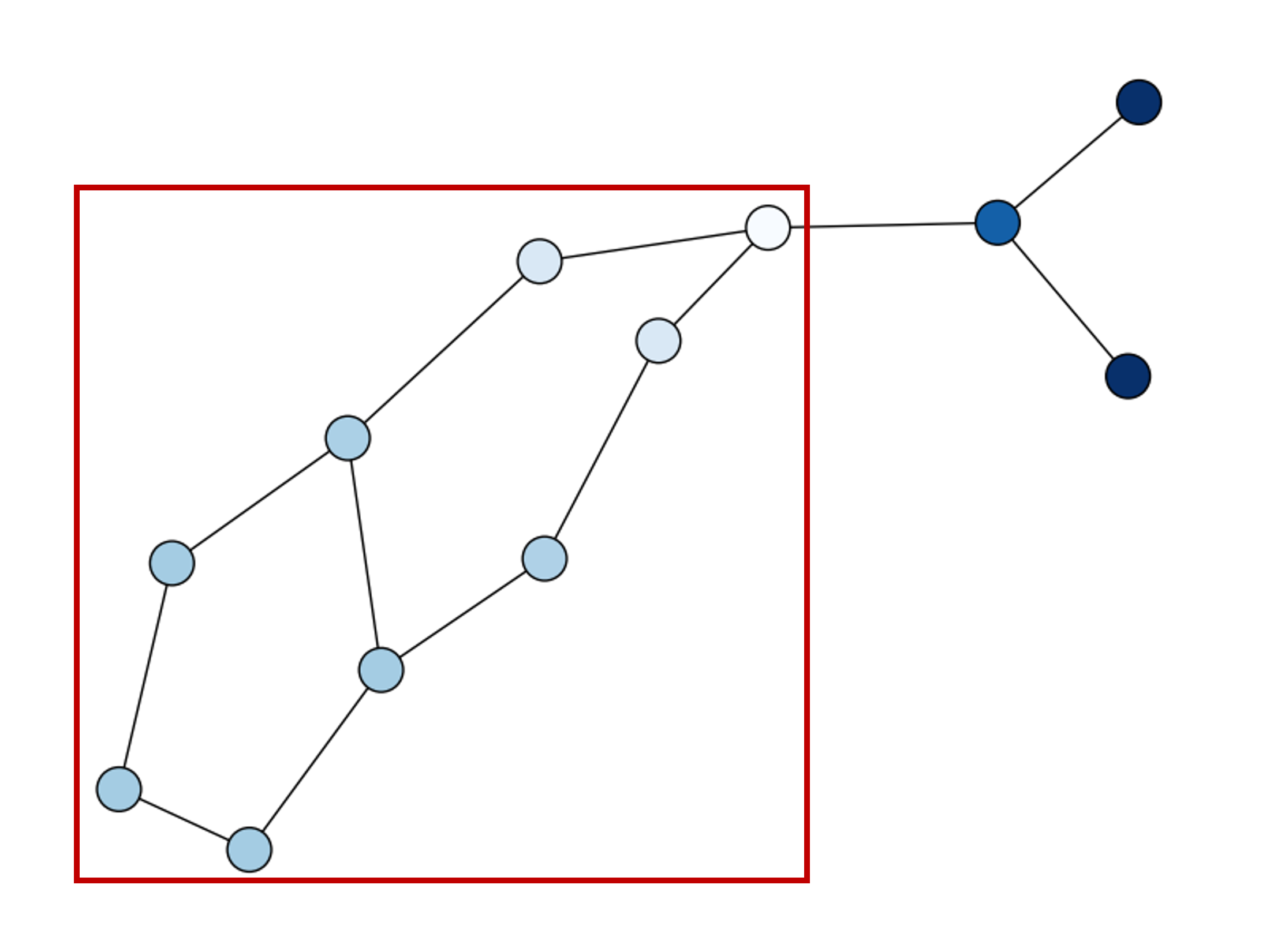} 
& \includegraphics[width=0.36\textwidth]{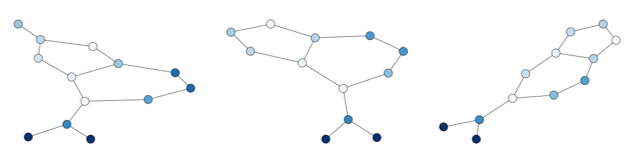}  \\
\hline
\end{tabular}
\caption{Visualizations of explanations generated by SIGNET, GLADPro, and ProtoGLAD on the MUTAG dataset.
For each method, we show the top-2 ranked anomalous graphs detected, together with a normal prototype graph and additional normal reference graphs.
For ProtoGLAD, the prototype graphs and their clusters are directly discovered from the dataset.
For GLADPro, whose prototypes are learnable latent vectors, we retrieve the closest normal graph in the embedding space as a reference prototype and show its nearest normal neighbors.
SIGNET does not explicitly identify normal prototypes or clusters, the prototype and normal reference examples shown for SIGNET are obtained by applying ProtoGLAD to the detected anomalous graphs.
}
\label{tab:comparison}
\end{table*}

We then evaluate our method's interpretability on real-world datasets.
We argue that in real-world datasets, graphs are not monolithic but often form multiple patterns (\textit{i.e.}, clusters). ProtoGLAD is designed to first discover these clusters and their corresponding centers, which serve as the prototypes.
Based on this, ProtoGLAD can provide contrastive explanations for its results.
For each detected anomaly, we first identify its nearest prototype and closest cluster. This enables a comparison between the anomalous graph and the nearest prototype. As illustrated in Table~\ref{tab:explainable}, the anomaly exhibits substructural deviations from its corresponding normal counterpart.

We also conduct a qualitative comparison with two explainable methods, SIGNET and GLADPro, on the MUTAG dataset as illustrated in Table~\ref{tab:comparison}.
\begin{itemize}
    \item \textbf{SIGNET} identifies an important ``bottleneck subgraph'' within the graph to indicate where an anomaly may occur.
    While it indicates where the anomaly is, SIGNET fails to provide a corresponding normal reference, making it difficult to verify whether the detected anomaly truly deviates from the majority.
    Notably, the top-1 ranked anomaly (the first row in Table~\ref{tab:comparison}) detected by SIGNET is identified as a normal graph by both GLADPro and ProtoGLAD. 
    To make a fair comparison, we use ProtoGLAD to obtain the normal prototype and cluster examples for the anomalous graphs detected by SIGNET.
    As illustrated, the top-1 anomalous graph detected by SIGNET is highly similar to its nearest normal prototype as well as to other graphs within the corresponding prototype cluster, which contradicts the fundamental definition of anomaly.
    This also highlights the risk of misinterpretation when the explanation lack a contrastive normal reference.
    
    \item \textbf{GLADPro} learns global prototypes. However, as its prototypes are abstract latent vectors, it can only visualize them indirectly by retrieving a set of Top-K similar graphs from the dataset.
    While this provides a general sense of normal patterns, it forces user to manually summarize normality from a diverse set of retrieved graphs.
    
    \item \textbf{ProtoGLAD} provides direct, contrastive explanations for its detection results.
    Not only does ProtoGLAD highlight the anomalous substructure, but it also finds the specific real normal prototype that the anomaly should have resembled. In contrast to the baselines, node colors in ProtoGLAD describe the \textit{similarity} to the prototype: darker nodes indicate high similarity (consistency with normality), while \textbf{lighter nodes highlight the anomalous deviations} (low similarity).
\end{itemize}

We provide a further analysis of the quality of the prototypes discovered by GLADPro and ProtoGLAD in Table~\ref{tab:comparison} and Table~\ref{tab:proto_compare}.
A critical observation from Table~\ref{tab:comparison} is that GLADPro incorrectly flags a ground-truth normal graph as anomalous.
We attribute this failure to the inherent instability of its prototype learning mechanism. Since GLADPro models prototypes as learnable vectors in a continuous latent space, these vectors may drift into sparse regions that do not correspond to any dense normal distribution.
This is evidenced in Table~\ref{tab:proto_compare}, where the second prototype retrieved by GLADPro is structurally similar to an anomalous pattern rather than a representative normal graph. Using such a reference inevitably leads to erroneous detection.
In contrast, ProtoGLAD explicitly identifies four concrete prototypes. As illustrated, ProtoGLAD not only captures the dominant normal pattern identified by GLADPro but also successfully discovers others that GLADPro fails to find.
This confirms that the data-dependent approach of ProtoGLAD ensures both the \textit{validity} and the \textit{coverage} of the normal reference set.

\begin{table}[tb]
\centering
\begin{tabular}{c|c}
\hline
\textbf{Method} & \textbf{Prototype} \\
\hline
GLADPro & \includegraphics[width=0.3\textwidth]{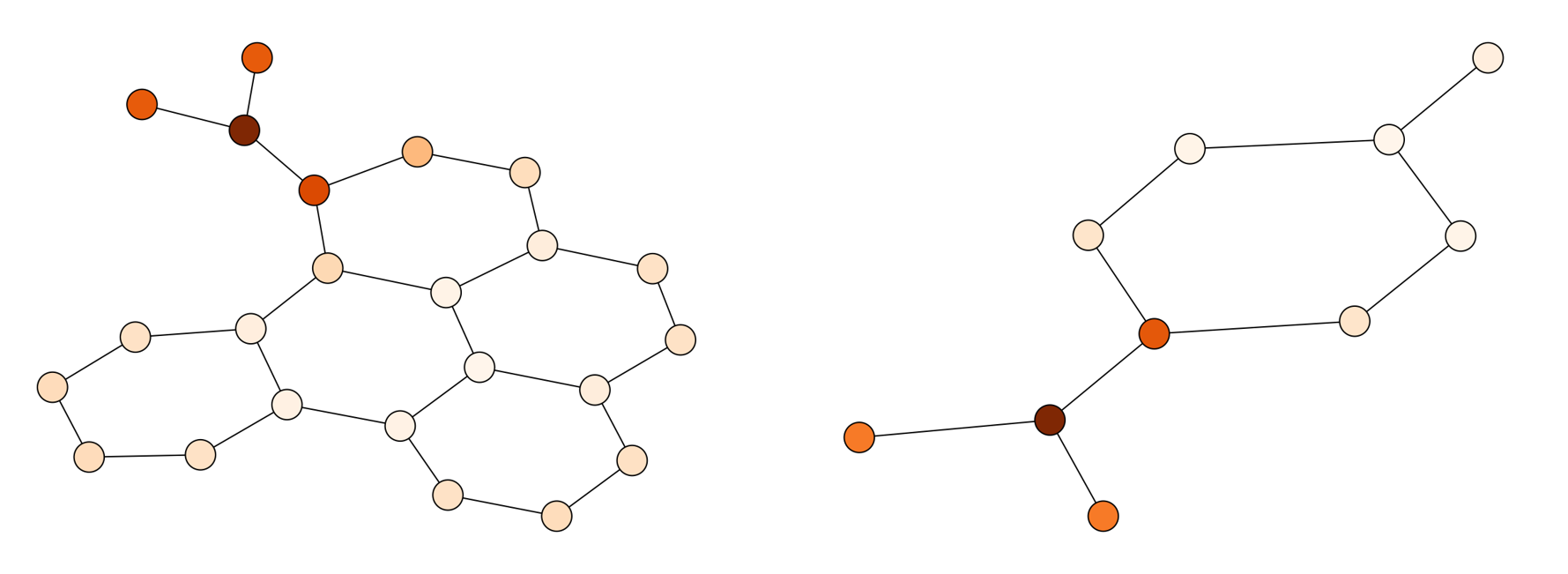} \\
\hline
\multirow{2}{*}{ProtoGLAD} & \includegraphics[width=0.3\textwidth]{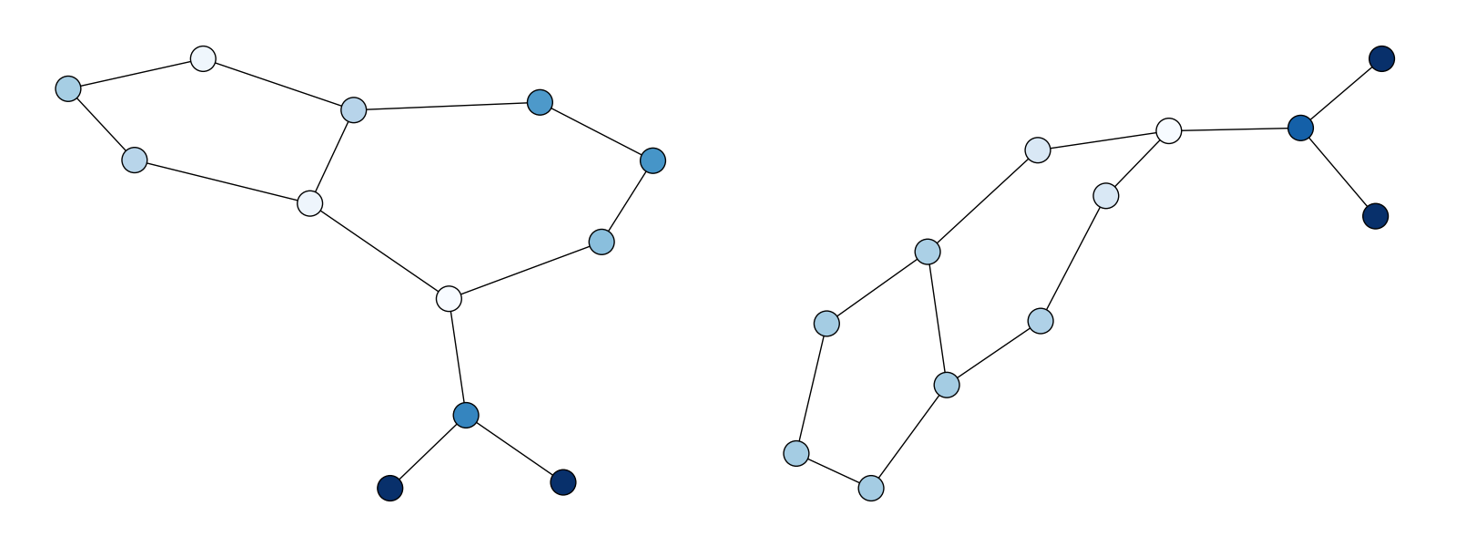} \\
 & \includegraphics[width=0.3\textwidth]{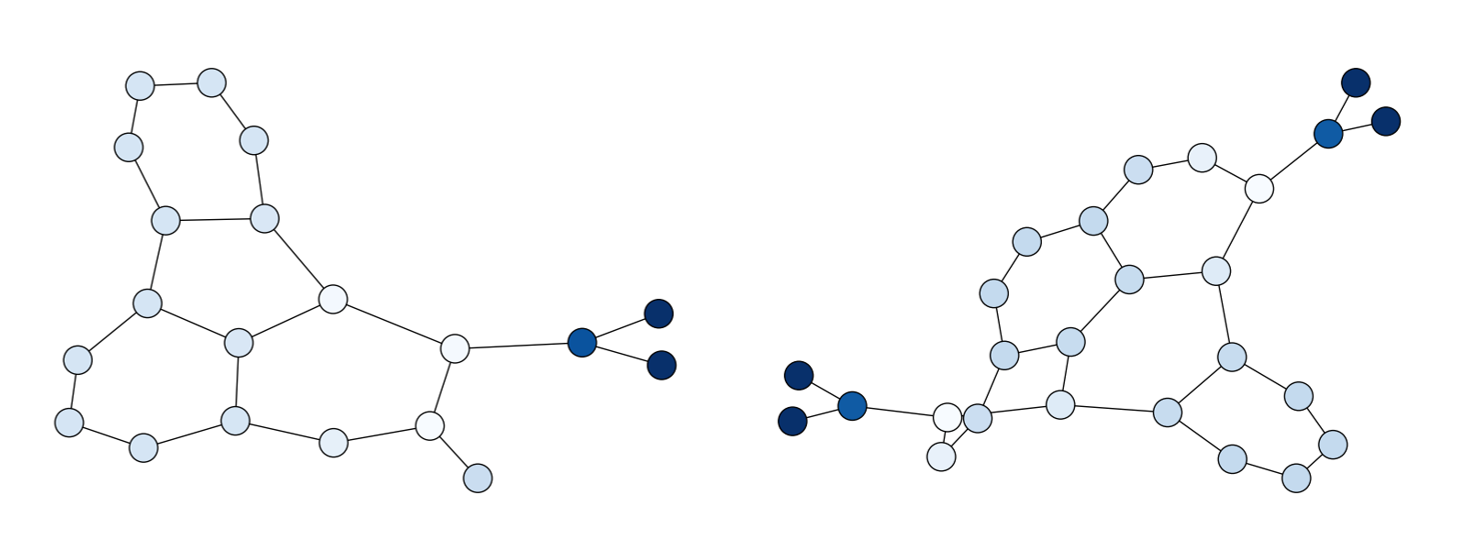} \\
\hline
\end{tabular}
\caption{Visualizations of prototype graphs identified by GLADPro and ProtoGLAD on the MUTAG dataset.}
\label{tab:proto_compare}
\end{table}

\section{Conclusion}

We present ProtoGLAD, an interpretable framework for graph-level anomaly detection that grounds its decisions in concrete normal graphs.
ProtoGLAD first discovers multiple modes of normality by clustering graphs with a point-set kernel and selecting representative normal prototypes, and then assigns anomaly scores based on each graph’s similarity to its closest normal cluster distribution.
Moreover, ProtoGLAD provides contrastive explanations: for a detected anomaly, it highlights the nodes/substructures that are least consistent with its corresponding prototype and the neighboring normal graphs, making the deviation from normal patterns explicit and verifiable.
Empirical evaluation results on both synthetic data with known anomalous motifs and multiple real-world benchmarks demonstrate that ProtoGLAD achieves strong detection performance while providing intuitive, prototype-based explanations.

\appendix

\bibliographystyle{named}
\bibliography{ijcai26}

\end{document}